# Interpretable Artificial Intelligence for Industrial Automation

[1] *A.G. Kolonin, ORCID: 0000-0003-4180-2870 <a.kolonin@g.nsu.ru>*
[2] *A.I. Dvortsevoy, ORCID: 0009-0004-3817-8813 <dvorcevoj@corp.nstu.ru>*
[1] *The Artificial Intelligence Research Center of Novosibirsk State University, 1, Pirogova st., Novosibirsk, 630090, Russia.*
[2] *Novosibirsk State Technical University 20, Prospekt K. Marksa, Novosibirsk, 630073, Russia*

**Abstract.** This article discusses an approach to developing artificial intelligence systems for industrial automation and the creation of automated process control systems—applicable to industry, municipal services, and infrastructure, with a particular focus on critical infrastructure. We propose an interpretable framework based on a hierarchy of finite-state machines that operate within the state-sequence space of a controlled system. Within this space, causal relationships—defined according to the Granger causality criterion—can be identified between state variables and control inputs; these relationships are represented as graphs of state transitions and their corresponding sequences. The feasibility of this approach is demonstrated through a "Smart Home" simulation using the Home Assistant platform, and real-world field data from a heat and water supply facility. This approach is presented as a foundation for building trustworthy artificial intelligence systems.

**Keywords:** industrial automation, automated process control system, interpretable artificial intelligence, trusted artificial intelligence, finite-state machine, state space, state-transition graph, Granger causality.

**Acknowledgements.** The results were obtained using resources of the projects «Aigents», «Unisoft Plus», and «TechnoAnalytica».

## 1. Introduction

The development of industrial automation systems is characterized by a consistent transition from rigidly defined control algorithms to adaptive control models capable of operating under conditions of uncertainty, incomplete observed data, and changing external disturbances. The application of machine learning methods to predictive equipment maintenance, process fault diagnostics, energy consumption optimization, and utility infrastructure dispatching allows for improved control of facilities for which the development of precise analytical models is difficult or economically unjustified. These tasks are particularly important for critical infrastructure facilities —heat and water supply systems, energy supply, transportation, and municipal services—where control errors directly impact the safety and continuity of services.

At the same time, the widespread adoption of models with high expressive power (deep neural networks, decision tree ensembles, transformer architectures) is accompanied by an increasing complexity of their analysis and verification. The emerging contradiction between forecast accuracy and the transparency of the decision-making mechanism is described in the literature as the problem of interpretability of artificial intelligence (AI) [1]. Explanatory intelligence methods, including those based on the assessment of the contribution of features [2], allow us to characterize the associative dependencies recorded by the model; however, they do not reveal the structural organization of the control process and do not guarantee either the correctness of the explanation in relation to the internal implementation of the model or its correspondence to the physics of the object. For areas where the justification of each control action is required, this level of justification is insufficient [3].

An additional constraint is the regulatory context. Functional safety standards (IEC 61508), risk management requirements for the application of artificial intelligence systems (ISO/IEC 23894), and European AI regulations require traceability of decisions, the possibility of their independent audit, and the preservation of the human control role. It follows that interpretability in industrial applications cannot be considered an additional characteristic of the user interface: it must be a property of the knowledge representation model itself and the method for generating control actions. In other words, “trusted” AI-based solutions must ensure not only “local interpretability,” allowing for the explanation of specific decisions post-factum, but also “global interpretability,” enabling audit and expert verification of the model itself before its industrial application [4].

A promising approach to ensuring trust is implemented in classes of models with intrinsically interpretable semantics. These include finite-state machines (FSMs), transition graphs, models based on finite state sequences, and formal behavior specifications. Such models allow for exhaustive computational tracing, invariant checking, scenario replay, and undesirable state reachability analysis. However, their traditional use in automated process control systems (APCS) involves the manual formalization of expert knowledge, making the model construction process labor-intensive and poorly scalable to objects with a large number of state variables.

A promising approach to resolving this contradiction is the automatic extraction of representational structures from observed data while preserving formal semantics. In this regard, the analysis of time series of state variables and control actions using Granger causality testing [5] is of particular interest. It should be emphasized that Granger causality expresses not a philosophical causal relationship, but the predictive contribution of one signal to the prediction of another, given that other information is taken into account. Therefore, the detection of such relationships in observed data should be accompanied by testing for spurious associations caused by hidden variables, common external disturbances, and delays, as well as—if possible—interventional confirmation in a simulation environment.

This paper proposes an approach to constructing an interpretable artificial intelligence system for industrial automation, in which the modeling object is not a single observation, but a sequence of states of the controlled system. This representation can be implemented as a hierarchy of finite automata operating in a space of state sequences; relationships between state variables and control actions are identified based on the Granger causality criterion and are recorded in the form of transition graphs. The hierarchical organization allows for the distinction between levels of local behavior, coordinated behavior of subsystems, and target control of the system as a whole, ensuring the alignment of the formal verifiability of lower levels with the adaptability of upper levels.

The fundamental feasibility of the proposed approach is being tested using a "Smart Home" mockup built on the Home Assistant software platform, which simulates the distributed interaction of sensors and actuators. It is also being tested using real field data from a heat and water supply facility, containing gaps, measurement noise, and sampling irregularities. A combined analysis of these environments can help separate the method's data-driven properties from its structural limitations.

The main contribution of the work is: (1) defining a procedure for constructing a hierarchy of finite automata based on the identified causal relationships; (2) demonstrating the applicability of the approach to simulated and real industrial data. The combination of these properties—solution traceability to a specific sequence of transitions, reproducibility, the ability to verify invariants, and preservation of the human control loop—is considered as the basis for building trusted artificial intelligence systems.

The article is structured as follows. Section 2 presents the rationale for the approach to interpretable modeling of dynamic objects and formulates the requirements. Section 3 describes the approach to constructing a hierarchy of finite state machines and the procedure for identifying causal relationships. Section 4 is devoted to experimental testing of the proposed approach for two environments—a smart home and a heating and water supply facility. Section 5 discusses the scope of applicability and potential limitations, and Section 6 presents the conclusion and directions for further research.

## 2. Justification of the approach and formulation of requirements

A promising formulation of the problem, from our point of view, is presented in the work, where the behavior of the mental system of a living organism or an artificial control system of a certain technological object can be formalized in a state space, including sensations (signals from the field), actions (control actions), and sensations (target parameters) [6]. In this case, the model of the organism's behavior or the control of a technological process can be considered as a graph of transitions between previous sequences of states of new states, where each transition can be characterized by its probability, its usefulness in terms of needs or the achievement of the system's target parameters, and a set of actions necessary to implement this transition. In essence, such a model can be described in terms of finite automata theory [7, 8].

The available experience shows that the given approach can be successfully applied, among other things, to the problems of reinforcement learning, including those with delayed reinforcement [9]. According to the computational model used in previous works [6,9], as shown in Fig. 1, the agent records states in the form of vectors of state variables $s_t$ based on interaction with the external environment and its internal processes at each moment of time $t$ , where $s_t$ consists of measurements of sensors $f_t$, the degree of satisfaction of internal needs or achievement of target parameters $y_t$, and the recording of completed actions or implemented control actions $a_t$. During interaction with the operating environment, the agent forms episodic memory in the form of a sequence of state vectors as a sequence $T$ of states , on the basis of which models of interaction with the environment can be formed in the form of transition graphs from sequences of states to new states $\{s_\tau\}_{\tau\in\{-T,-T+1,\dots,-1\}} \Rightarrow s_{\tau=0}$, where $\tau$ is the relative time for each such transition from the final state to one of the previous ones. Each transition from a sequence of states to a new state can be characterized by a utility value in the model formed during training. $U$ , characterizing the degree of satisfaction of the agent's needs or the degree of achievement of its current target parameters, as well as the probability of transition to a new state $P$ , or the counter of the number of data transitions $C$ , which allows this probability to be calculated.

The learning function $L$ allows us to calculate the incremental change in utility $U$ during transitions between sequences of states at time $t-1$ to new ones at time $t$ according to the following generalized formula.

$$dU(\{s_\tau\}_{\tau\in\{-T,-T+1,\dots,-1\}}, s_{\tau=0}) = L(x\,(y_t - y_{t-1}), E(a_{t-1}))$$

In the formula above , $x$ is the vector of importance or priority of current needs or target parameters, $y_t - y_{t-1}$ is the vector of satisfaction of the system's needs or its proximity to the specified target parameters, which can be interpreted as a "reward" from the standpoint of reinforcement learning, or positive or negative feedback in the context of [6,9]. $E(a_{t-1})$ here determines the measure of energy or other resource expenditure on completing a given transition, if maximization of the energy or generally resource (e.g., economic) efficiency of the system's behavior is assumed.

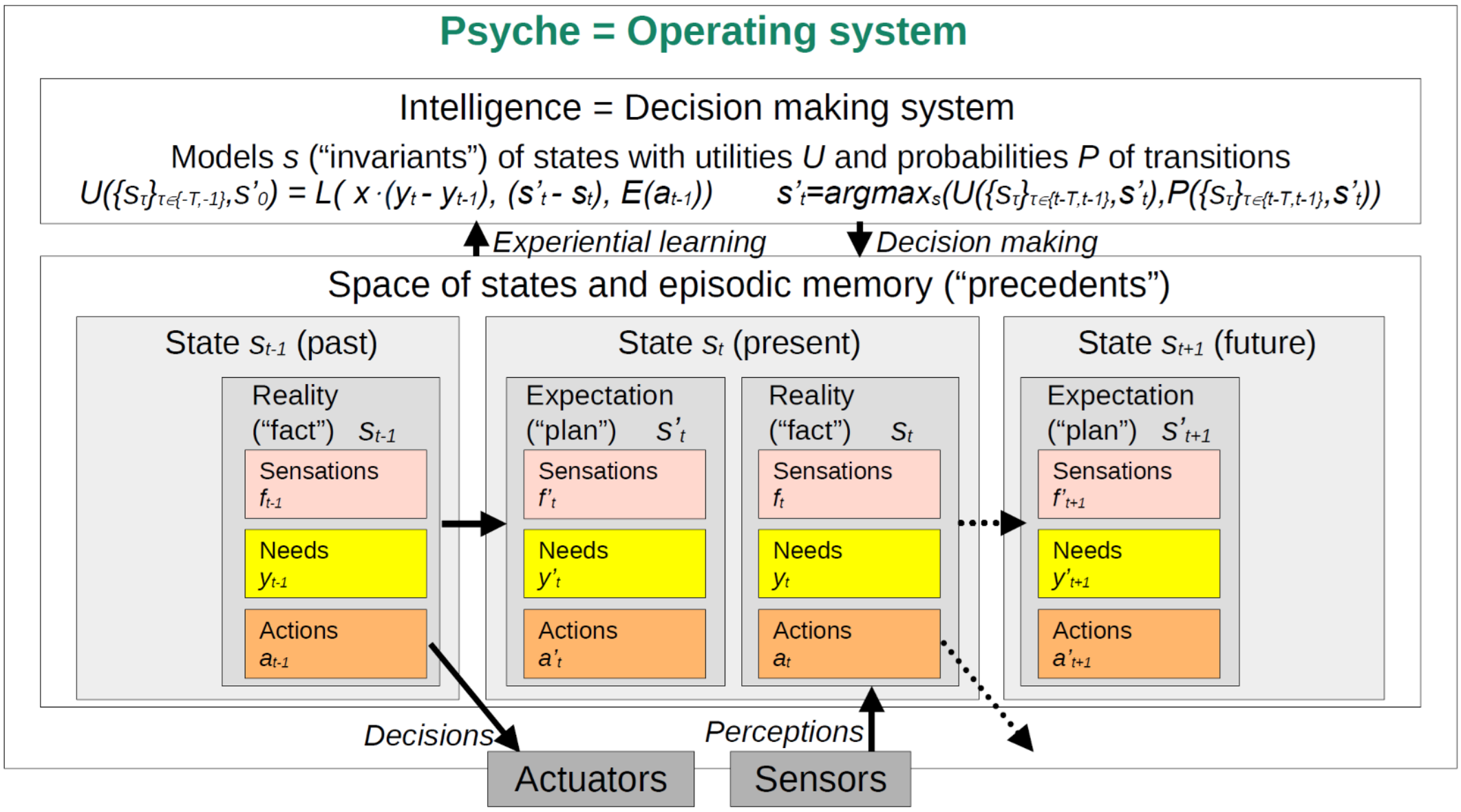

*Fig. 1. Illustration of the decision making process in the state of spaces based on state transition graphs stored in the episodic memory of an agent, where experienced states s are agent’s memories kept as precedents linked by transitions, associated with per-transition utility and evidence counts while predicted states s' are expectations that an agent experiences at any given moment while making a decision or calmly observing an environment.*

The decision function for such an agent can be based on maximizing the combination of the utility U of a transition from the sequence of states $\{s_\tau\}_{\tau\in\{-T,-T+1,\ldots,-1\}}$ at time $t$-$1$ to a new state $s_t$ at time $t$ and the probability P of this transition, calculated based on the distribution of known values of transition counters C for all known transitions from the original sequence of states.

$$s'_t = \operatorname{argmax}_s \left(U(\{s_\tau\}_{\tau\in\{-T,-T+1,\ldots,-1\}}, s_{\tau=0}), P(\{s_\tau\}_{\tau\in\{-T,-T+1,\ldots,-1\}}, s_{\tau=0})\right)$$

While this approach can be implemented within the paradigm of finite automata theory [7,8], a well-known problem with its application is the combinatorial explosion of states in flat automata when modeling complex systems [10,11], which in our case is aggravated by the fact that the state of the automaton is actually determined by an entire sequence of states $s$ over a certain time interval $T$. An effective solution to this problem requires both increasing the discreteness of the values of each state variable and identifying segments or hierarchies in the homogeneous state space s corresponding to various subsystems of the agent itself and the environment it controls, with reduced dimensionality for each such segment or section of the hierarchy. This should allow both to reduce the number of possible states of each individual variable and to minimize the number of variables in each individual state flow corresponding to an individual finite automaton in the hierarchy of such automata.

## 3. Approaches to model construction

The problem of a combinatorial explosion of the number of states in flat automata [7,8] is solved by both increasing the discreteness of each state variable and decreasing the number of state variables—dividing the overall variable space into isolated groups (corresponding to individual physical subsystems of the control object) or intersecting groups (in the case where some environmental variables influence several groups or are subject to the influence of several groups). The corresponding isolated groups of variables can be brought into correspondence with individual finite automata, and the control model of the object (the behavior model of the control agent) can be reduced to a superposition or hierarchy of such automata.

## 3.1 Signal quantization

Signal quantization involves assessing the nature of the distribution of possible values of a variable corresponding to a signal, with the aim of assessing the nature of the distribution of values, identifying the presence of several modalities that potentially correspond to different states or modes of a system, the parameter of which is determined by a signal whose values are represented by the corresponding variable.

In our previous work [12], the solution to this problem is proposed by a combination of the dip test metric (Hartigan's statistical test) [13], which determines the degree of unimodality of the distribution, and the Silhouette Coefficient [ 14], which allows us to evaluate the optimality of clustering for a given number of clusters in a multidimensional space (in our case, for clustering using the K-means method [12] in a one-dimensional space).

In this study, we used the same approach: we assessed the possibility of identifying more than one modality using the " dip test" metric and determined the optimal number of modalities as the number of clusters maximizing the "silhouette" coefficient for one-dimensional K-means clustering. Accordingly, quantization was accomplished by replacing the original signal values with their values corresponding to the "centroid" of the closest K -means "cluster" with the optimal number of clusters K corresponding to the number of signal modes or regimes.

## 3.2 Identification of group states

The problem of identifying groups of signals representing parameters of an environment or system that are interconnected due to their intrinsic nature was considered by us in previous works on causal and predictive analytics as applied to the analysis of financial time series [15,16]. For this purpose, we used an analysis of the values of the Pearson correlation coefficient between the values of the time series, obtaining cross-correlation matrices with values indicating the degree of interdependence of the signals, either direct (positive values) or inverse (negative values). Also, to identify potential cause-and-effect relationships on different time horizons based on the principle of Granger causality [5], we analyzed the values of the Pearson correlation coefficients between different time series with different time shifts, determining those shifts that maximize the absolute value of the correlation coefficient (negative or positive relationship). The potential causal relationships thus identified were used to construct predictive models based, for each target signal, on the signals with the greatest identified causal relationship to the target.

In our present study, the goal was to identify distinct groups of signals interconnected across different time horizons. Therefore, we explored the possibility of clustering signals from the entire control object into groups corresponding to individual subsystems. For these groups, it was then possible to construct predictive models in the state space of these subsystems, with the goal of controlling the system as a whole based on a hierarchy of finite-state machines reflecting the hierarchy of the control object's subsystems. In the study, we found that many signals are characterized by significant changes in signal level over the observation period, that is, low-frequency changes (on the scale of months and days) are significantly more significant in terms of signal amplitude than high-frequency changes on the scale of hours and minutes. In this regard, the analysis of the mutual correlation between signals both “in the moment” (without a time shift) and with time shifts (for the purpose of causal analytics according to Granger) was carried out both on the original signals (with the identification of low-frequency connections on the scales of months and days), and on the high-frequency data (hours and minutes) reduced to a common level, obtained by normalizing the signals $S_t = S_t / max(abs(S_t))$ and subtracting the low-frequency "trend" $S^*_t = S_t - rollingmean_{window=7}(S_t)$, obtained by smoothing in a sliding window (for example, for the second experimental environment, the window width for hourly measurements was empirically determined as 6 hours, or seven measurement points) .

An analysis of the distribution of the values of the cross-correlation matrices made it possible to identify, by clustering, groups of strongly interconnected signals with cross-correlation coefficients (for the second experimental environment, above 0.9), and an analysis of Granger causality made it possible to identify pairs of signals that are potentially related by cause-and-effect relationships with different time horizons (for the second experimental environment, 1 hour and 6 hours). The paired relationships identified as described above allowed us to construct groups of interconnected signals across different time horizons. For example, for the second experimental environment, we could identify groups: a) related "at the moment" (based on unshifted correlation coefficients), b) related by cause-and-effect relationships over an hourly interval, and c) related by cause-and-effect relationships over a six-hour interval.

## *4. Experimental results*

### 4.1 «Smart Home»

The first environment studied was a simulation of a smart home automation project, built using the Home Assistant platform ( https://www.home-assistant.io/ ). This platform was used to deploy a virtual configuration of two rooms with motion sensors and lighting equipment (lamps), as shown in Figure 2.

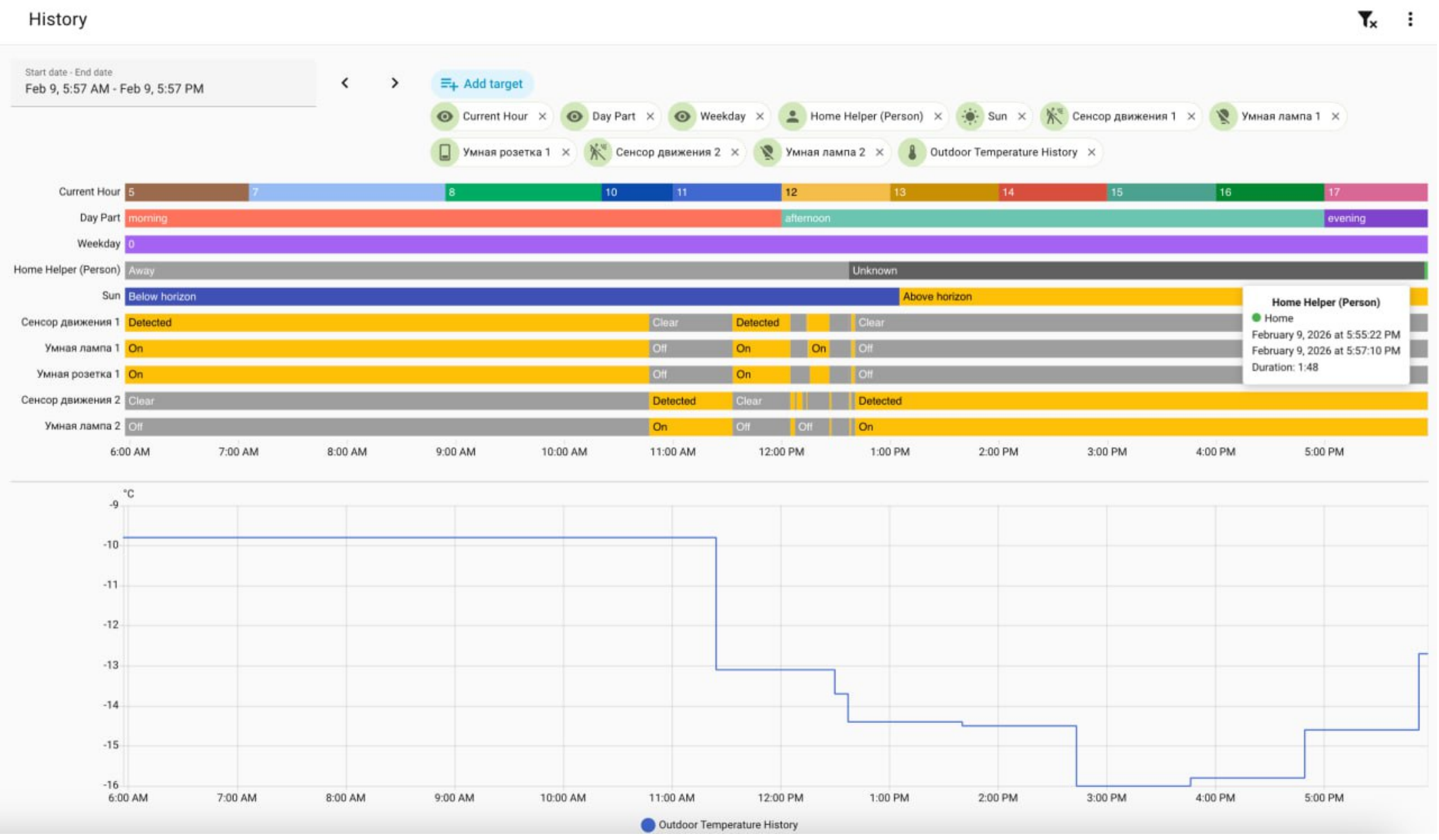

*Fig. 2. Visualization of the temporal dynamics of measured and control signals for a virtual configuration of two rooms equipped with motion sensors and lighting fixtures (lamps), implemented on the Home Assistant platform.*

The study aimed to automatically generate a state transition model that would enable adaptive adjustment of the automatic switching on and off of lighting equipment in real time based on the accumulation of statistics on the behavior of a virtual home occupant, as recorded by motion sensors and the switching on and off of lighting equipment.

The study identified distinct patterns or states of combined motion sensor activation and specific combinations of enabled lighting fixtures. With sufficient statistical prominence, these patterns could predict transitions such that partial activation of a given state (for example, the activation of a motion sensor in one room along with the activation of one of the lamps in the same room) could automatically trigger control signals that bring the current state to a complete combination with the statistically expected state (for example, the activation of missing lamps, which are typically on in this situation).

## 4.2 Heat and water supply system of the municipal economy

The study involved a section of a dual-circuit heating system with intermediate heat exchange points as shown in Fig. 3.

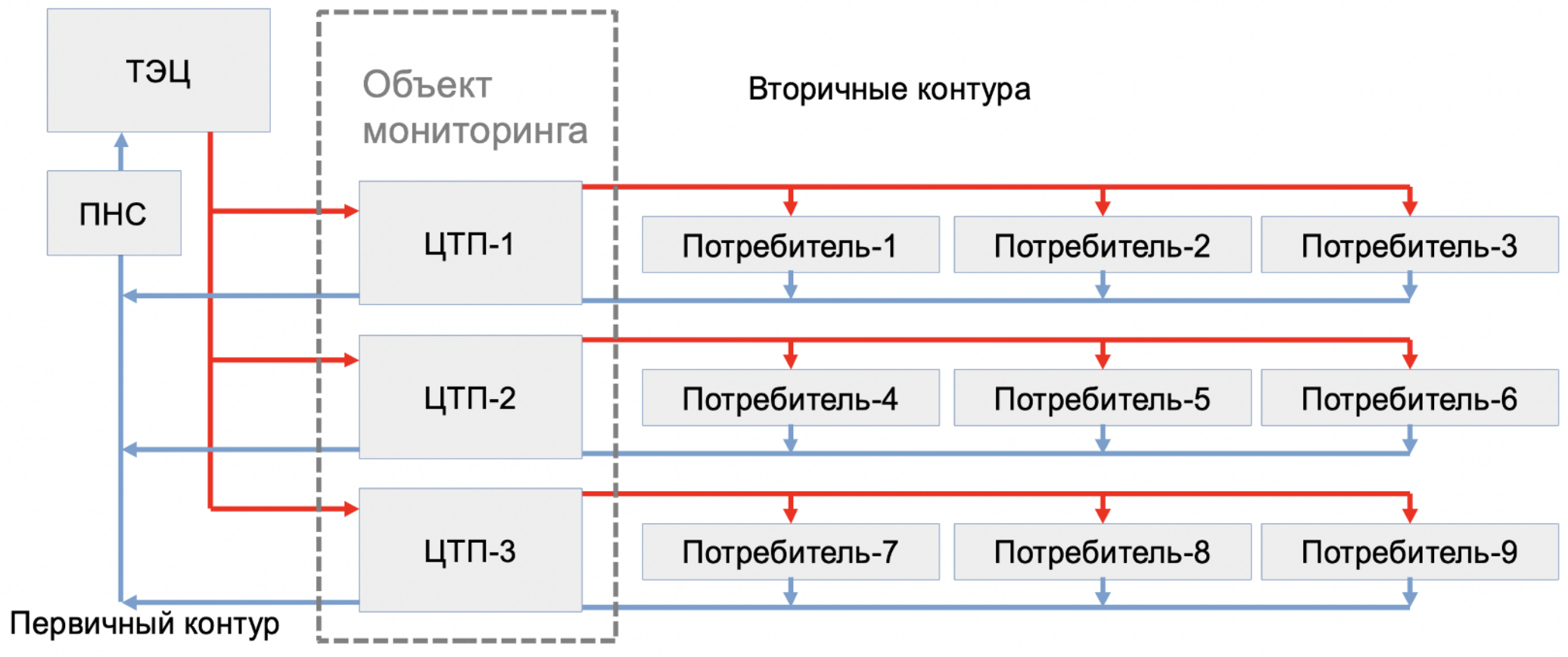

*Fig. 3. Schematic of the two-circuit district heating system with three central heating substations, the return-line booster station and the DHW circuit (consumers are indicated for illustrative purposes).*

The primary circuit consists of main lines from the combined heat and power plant CHP (ТЭЦ) through three central heating units CHP-1, CHP-2, CHP-3 (ЦТП-1, ЦТП-2, ЦТП-3) and a return pipeline booster pump station BPS (ПНС) with the coolant returning to the CHP. Consumers in the secondary circuit are connected to each CPH. The hot water supply DHW (ГВС) circuit includes DHW heaters in the CPH; cold water is supplied to the CPH from the city water supply. Primary circuit makeup is provided at the CPH and measured by an M94 flow meter. A significant feature of the facility is the geodetic difference: the CHP is located on a hill (plateau), while the consumers are in a lowland. The difference in elevation necessitates the presence of a BPS on the return pipeline.

The measurement boundaries (Fig. 3, Table 1) cover the primary circuit from the central heating station inlets to the pumping station (PS), the hot water supply circuits, and the cold water inlets. Heat output from the cogeneration plant, parameters on the pressure side of the PS, and subscriber parameters are not included in the data. This results in an incomplete heat balance for the plateau-lowland main line: the amount of heat output by the source cannot be compared with the sum Q for the central heating station. The balance is verified only for individual nodes — by the discrepancy between the primary circuit flow rates and the ratio of the hot water supply circuit flow rates. The data represents a telemetry archive from three central heating stations for the period from 10.01.2021 to 09.12.2024. The original sample is irregular; for analysis, it is converted to a uniform hourly grid by averaging, while intervals without data retain the omission flag. The composition and characteristics of the channels are presented in Table. 1. The set of channels varies across different central heating stations: some channels are missing from the archives of individual nodes. When constructing the connection graph, such a channel was excluded from the corresponding component rather than reconstructed from adjacent nodes: reconstruction would introduce a connection into the analysis that is not present in the data. Also, to obtain the results below, records outside the period from December 13, 2022, to April 16, 2023, were excluded due to the presence of significant sections with missing or constant values for certain signals. Ultimately, the presented results were obtained from 2,106 measurements for each of the 55 signals.

*Table 1. Measurement channels for the object.*

| Designation | Physical meaning | Circuit | Units of measurement |
|---|---|---|---|
| T1 | Temperature of the coolant in the supply pipeline | primary | °C |
| T2 | Temperature of the coolant in the return pipeline | primary | °C |
| P1 | Pressure in the supply pipeline | primary | kgf/cm² |
| P2 | Return line pressure | primary | kgf/cm² |
| M1 | Mass flow rate at the inlet of the central heating plant | primary | t/h |
| M2 | Mass flow rate at the outlet of the central heating station | primary | t/h |
| M94 | Make-up flow rate | primary | t/h |
| T3 | Hot water temperature in the supply pipe | hot water supply | °C |
| T4 | Hot water temperature in the return pipe | hot water supply | °C |
| P3 | Pressure in the DHW supply pipeline | hot water supply | kgf/cm² |
| P4 | Pressure in the return pipeline of hot water supply | hot water supply | kgf/cm² |
| M3 | Mass flow rate of hot water at the inlet | hot water supply | t/h |
| M4 | Mass flow rate of hot water at the outlet | hot water supply | t/h |
| $T_{hvs}$ | Temperature of cold water at the central heating station inlet | cold water supply | °C |
| $P_{hvs}$ | Cold water pressure at the central heating station inlet | cold water supply | kgf/cm² |
| $V_{hvs}$ | Cold water flow rate | cold water supply | m³/h |
| $V_{hvsb}$ | Bypass flow rate | cold water supply | m³/h |
| Q | Thermal energy of heating and ventilation | primary | Gcal/h |
| $Q_{gvs}$ | Thermal energy of hot water supply | hot water supply | Gcal/h |
| $T_{nv}$ | Outdoor air temperature | external | °C |

During the quantization studies, two types of signals were identified: those characterized by a near-normal distribution of values ( dip test > 0.65 and Silhouette Coefficient < 0.8) and those characterized by a near-multimodal distribution ( dip test <= 0.65 and Silhouette Coefficient >= 0.8). For the latter signals, the typical number of modes identified by the analysis was 2-3. To solve the subsequent problem of identifying final states for constructing predictive models based on finite state machines, all signals were quantized according to the optimal number of modes from 2 to 3, depending on the maximum value of the Silhouette Coefficient for a given number of modes, with each signal value replaced by the closest possible value of the corresponding mode, calculated as the mean for the mode (corresponding to the centroid of the one-dimensional clustering). An example of quantized signals is shown in Fig. 4.

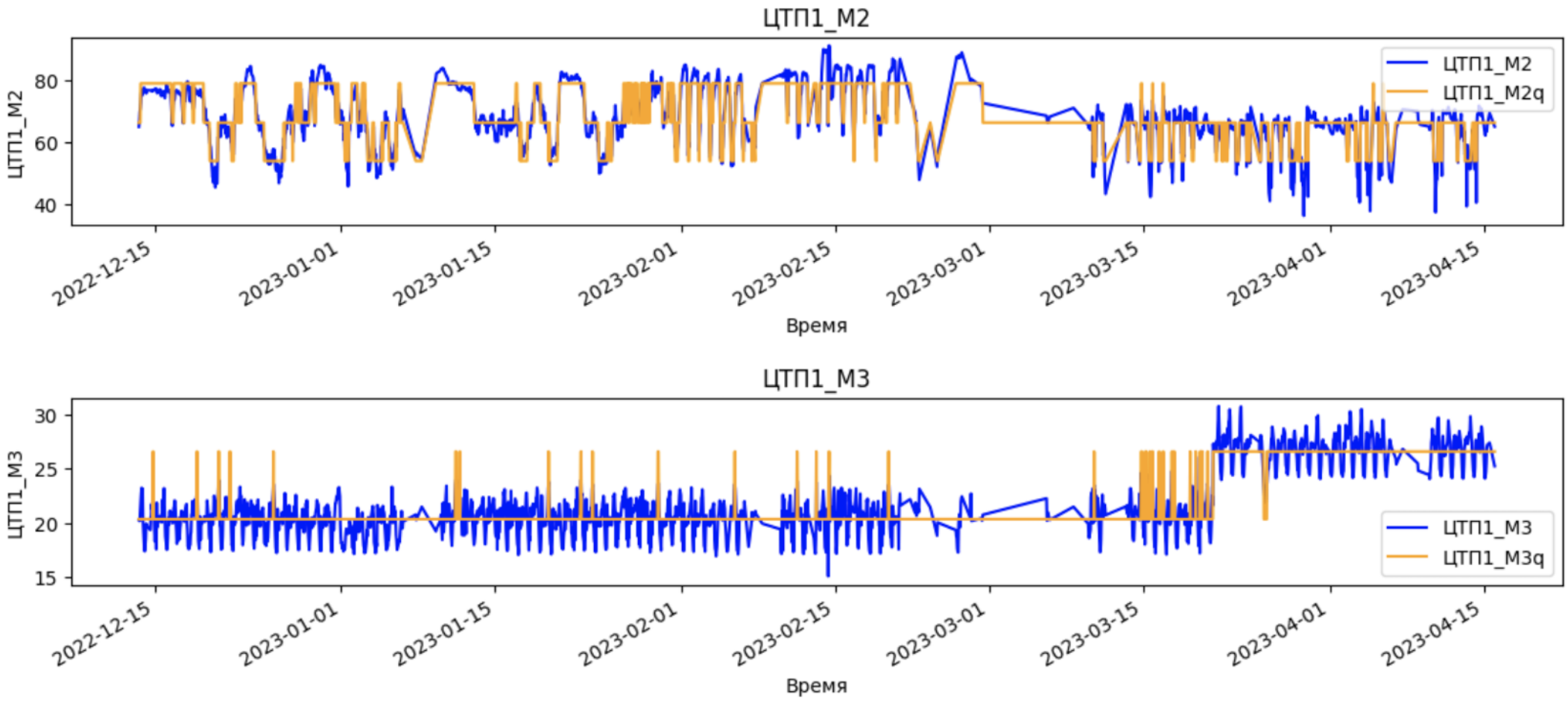

*Fig. 4. Quantization results for two signals (blue — original signal, orange — quantized signal).*

A real-time correlation analysis of the original signals and signals with their low-frequency "trend" removed allowed us to identify two types of signal groups. In the first case, these were groups of signals linked by low-frequency correlation (at the daily and monthly levels), sharing common seasonal components, or relating to subsystems of the facility with common conditions in a given operating mode (e.g., a common emergency or a common normal mode). In the second case, groups of relatively high-frequency signals with their "trend" removed were identified, calculated as a 7-hour moving average. Figure 5 shows groups of original signals with a high degree of statistical connectivity—either direct or inverse correlation with an absolute value of the Pearson coefficient greater than 0.9.

The resulting graph is decomposed into components corresponding to physically distinguishable subsystems: the primary circuit of an individual central heating station (CHS)—the ratio of flow rates, pressure drops, and temperatures; the general temperature regime—the coordinated change in T1 at different CHSs with a transport delay; the DHW circuit—the balance of flow rates M3, $V_{hvs}$ , and thermal energy $Q_{gvs}$ ; the DHW hydraulics—the pressures P3, P4, and $P_{hvs}$ . Links between identical channels of different CHSs are preserved in the graph only in the presence of a nonzero lag consistent with the transport delay in the main line. With a zero lag, they are interpreted as a manifestation of a common external factor and are removed from the graph.

Granger causality was determined after subtracting the "trend" from the signal, which included a seasonal component within uniform mode windows, ensuring approximate stationarity of the series. The width of a 6-hour sliding window (7 measurement points) for calculating the subtracted trend was determined experimentally.

A search for time shifts yielding the maximum correlation coefficient was performed for each pair of signals with shifts from 1 to 24 hours. The results of clustering signals with identified potential cause-and-effect relationships are shown in Figure 6. Two classes of relationships were identified: relatively short-term relationships with a 1-hour shift and relatively long-term relationships with a 6-hour shift. This suggests the possibility of constructing interpretable predictive models for a number of parameters based on the identified hypothetical cause-and-effect relationships between recorded signals for 6-hour and 1-hour forecast horizons.

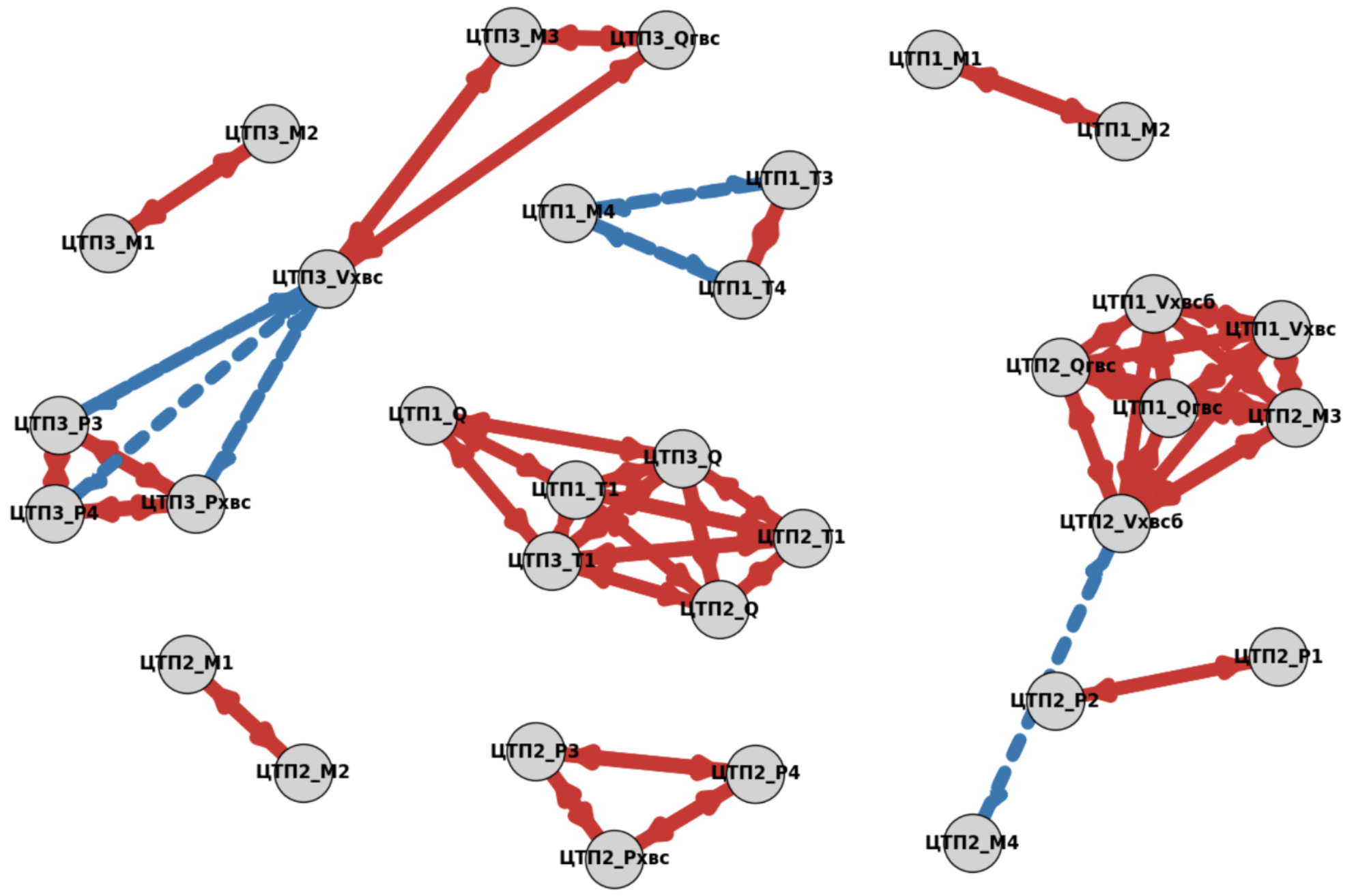

*Fig. 5. Signal connectivity graph based on positive (solid red arrows) and negative (dashed blue arrows) Pearson cross-correlation with absolute values greater than 0.9; the thickness of the arrows corresponds to the value of the correlation coefficient.*

A significant limitation of the method for this facility is that the heat supply loops are closed by controllers: the DHW temperature and pressure drop are maintained by automatic controllers, resulting in measured series generated in the "setpoint - control valve - flow rate - temperature" feedback loop. Under these conditions, the criterion identifies a directional contribution to predictability, rather than physical causality, and may indicate a direction opposite to the actual impact. The results are interpreted as "directional influence in the sense of predictability" and are formulated as such in the context of the study.

The robustness of the result to the seasonality removal method is separately tested. Relationships that persist across levels and residuals are considered physically justified; relationships that disappear when the seasonal component is removed are not considered. For example, in the preliminary analysis, the relationship between the return network water temperature of CHP-2 and the cold water temperature of CHP-1 was highly significant at the series levels, but disappeared after removing the seasonal component; this relationship is classified as an artifact of a common external factor. The only relationship between objects that is robust to all processing options is the directional influence of the network water temperature of CHP-3 on the temperature of CHP-2.

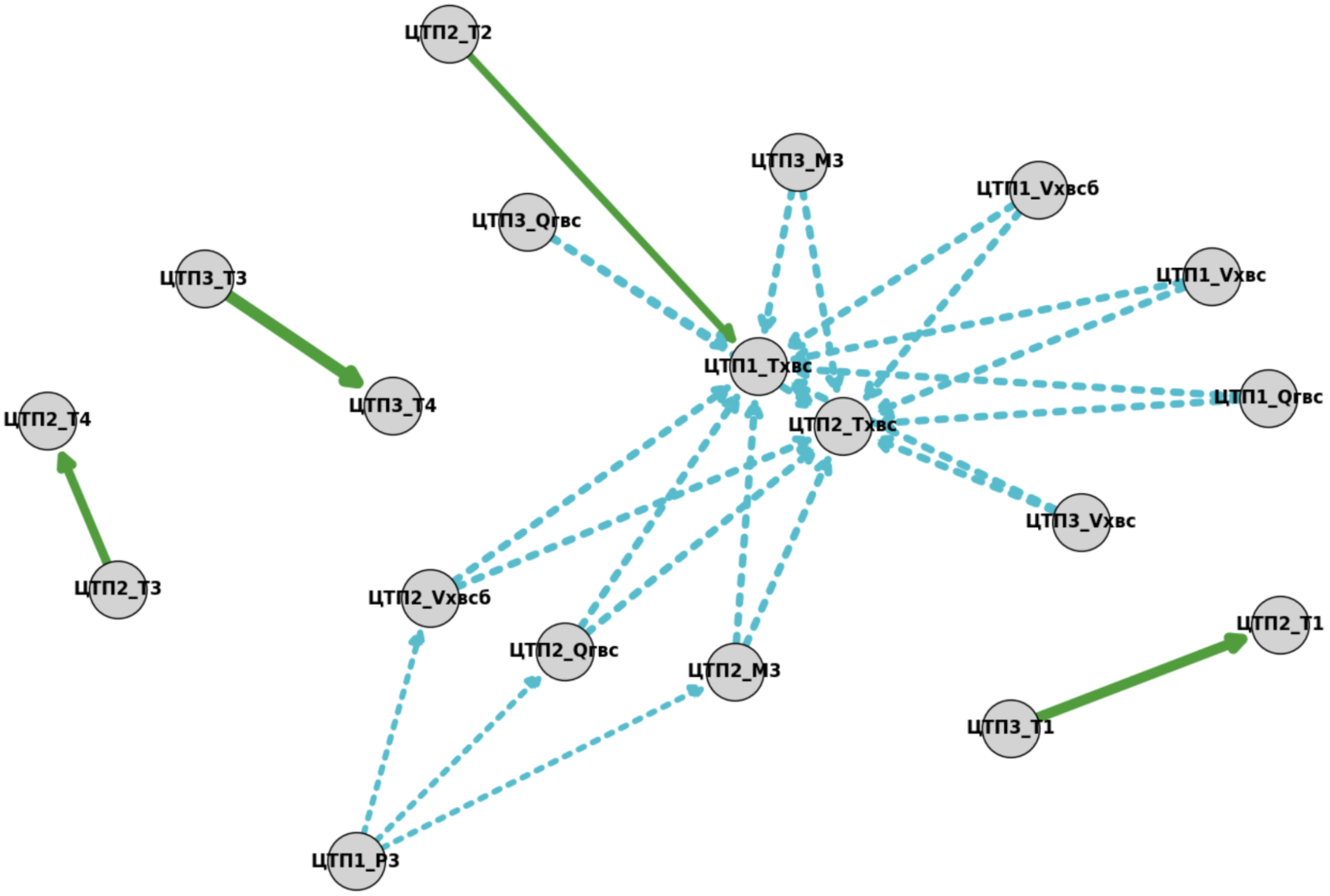

*Fig. 6. A graph of potential causal connectivity between signals (in the Granger sense), based on positive or negative cross-correlation at various time lags: solid green arrows represent a 1-hour lag, while dashed light-blue arrows represent a 6-hour lag; the thickness of the arrows corresponds to the magnitude of the correlation coefficient.*

An experiment to construct a finite-state machine model based on the obtained quantized signals was initially conducted without signal grouping, using all 55 signals for 2,106 measurements for each channel. In this case, only 1,968 unique states were detected, meaning that less than 10% of all states were reproducible only a few times during the study period, indicating the practical impossibility of constructing a model of the entire system for the controlled object.

Next, this experiment was repeated individually for each of the signal groups identified above on the basis of mutual correlation (Fig. 5), and in this case the number of unique states turned out to be three orders of magnitude smaller, as shown in Table 2. That is, for each of the identified groups, the overwhelming majority of the recorded discrete states for multichannel groups of quantized signals turned out to be represented by a multitude of episodes, which can make it possible to construct predictive models based on the theory of finite automata and transition graphs between sequences of states and individual states.

*Table 2. Number of detected discrete states, by groups of quantized channels.*

| State variables in a signal group | Number of states |
|---|---|
| ЦТП1_М1, ЦТП1_М2, ЦТП1_П2 | 15 |
| ЦТП1_М3, ЦТП1_М4, ЦТП1_Т3, ЦТП1_Т4 | 7 |
| ЦТП1_П3, ЦТП1_Qгвс, ЦТП1_Вхвс, ЦТП1_Вхвсб, ЦТП2_М3, ЦТП2_М4, ЦТП2_Qгвс, ЦТП2_Вхвсб, ЦТП3_М3, ЦТП3_П3, ЦТП3_П4, ЦТП3_Пхвс, ЦТП3_Qгвс, ЦТП3_Вхвс | 135 |
| ЦТП1_Q, ЦТП1_Т1, ЦТП1_Т2, ЦТП2_Q, ЦТП2_Т1, ЦТП3_Q, ЦТП3_Т1, ЦТП3_Т2 | 62 |
| ЦТП2_М1, ЦТП2_М2, ЦТП3_П1 | 13 |
| ЦТП2_П1, ЦТП2_П2 | 4 |
| ЦТП2_П3, ЦТП2_П4, ЦТП2_Пхвс | 3 |
| ЦТП3_М1, ЦТП3_М2 | 4 |

# *5. Areas of applicability and limitations of the approach*

## 5.1 Areas of applicability

The proposed approach to constructing interpretable artificial intelligence, based on a hierarchy of finite state machines and Granger causality analysis in the space of state sequences, opens up new possibilities for managing complex dynamic systems. Unlike statistical (neural network) "black-box" models, this method ensures formal traceability of decisions, which is critical for critical infrastructure facilities, such as housing and utilities systems.

For example, applying the described approach in the housing and utilities sector transforms the control system from reactive (responding to existing failures) to proactive and predictive. Moreover, rigorous mathematical formalization in the form of finite-state machines ensures that the complexity of the algorithms does not compromise the clarity and controllability of processes for the human operator.

One of the main advantages of the FSM hierarchy and cause-and-effect analysis is dimensionality reduction. In our study, decomposing 55 channels into eight groups of 2–14 signals (Table 2) reduces the amount of manual analysis: the number of distinguishable states per group does not exceed 135, compared to 1,968 in the original version. The constructed model allows us to identify and record the most energy-efficient operating modes in the transition graph, while rejecting suboptimal ones. Another important area of application is identifying causal relationships, which may allow the system to recommend smooth, timely, and predictable mode shifts rather than abrupt, discrete parameter changes. In our example, this should reduce hydraulic and thermal shocks in the network.

## 5.2 Limitations of the approach

The main limitations of the approach currently appear to be the lack of practical testing. Specifically, when testing the approach in a Smart Home environment, the environment was overly simplified, no actual signal grouping was performed, and the overall model was built for the entire facility over a limited time range. This cannot be considered convincing. Testing is needed on real physical configurations, with the construction of actual subsystem hierarchies within the stated approach.

With regard to the experiment described for the water supply facility, the validity of quantizing signals with a near-normal distribution, identifying only 2 or 3 discrete states, remains controversial. Overall, the resulting subsystems of final states for groups of signals require confirmation of their practical significance. This requires the construction of real predictive models and an assessment of their actual accuracy. The predictive power of the identified Granger causal relationships also requires separate verification, taking into account expert validation of all identified relationships.

## *6. Conclusion*

In the presented work, we demonstrated the potential for practical applicability of interpretable AI based on the theory of finite automata in the state space of environmental variables and transition graphs between sequences of discrete states and new states, based on the quantization of initial signals, with the construction of a hierarchy of state flows, corresponding hierarchies of corresponding finite automata - using correlation analysis of signals and analysis of potential causal relationships between signals according to Granger.

As we continue to develop our work, we plan to validate the proposed approach using more realistic simulations and real-world Smart Home installations, as well as using more comprehensive, voluminous, and high-quality data from housing and utilities and industrial facilities, including the full implementation and testing of predictive models based on the proposed approach.

## *Information about authors*

Anton Germanovich KOLONIN is a PhD in engineering (computer science), a leading researcher at the Artificial Intelligence Research Center, and a senior lecturer at the Faculty of Mathematics and Mathematics at Novosibirsk State University in Russia. He is the founder of the open-source software platform Aigents, a leading AI researcher at SingularityNET, and the creator of the AGI-in-Russian community. His research interests include strong artificial intelligence (AGI), natural language processing (NLP), blockchain architecture, social network analysis, knowledge extraction, machine learning, and industrial automation.

Alexander Igorevich DVORTSEVOY is a PhD in engineering, an associate professor at the Department of Thermal Power Plants at Novosibirsk State Technical University (NSTU NETI). He leads projects at Unisoft Plus (develop an IIoT platform covering 518 process lines at 198 enterprises for Fortune 500 companies) and is the scientific supervisor of TechnoAnalytica, a Skolkovo resident. His research interests include automation and digital control of thermal power facilities, power plant process control systems, digital twins, machine-learning-based monitoring and diagnostics of power equipment (district heating substations and power transformers), modelling and optimization of power unit operating modes, exergy analysis and excess fuel consumption, and LLM-based interpretation of process data.

# Интерпретируемый искусственный интеллект для промышленной автоматизации

[1] *А.Г. Колонин, ORCID: 0000-0003-4180-2870 <a.kolonin@g.nsu.ru>*
[2] *А.И. Дворцевой, ORCID: 0009-0004-3817-8813 <dvorcevoj@corp.nstu.ru>*

[1] *Исследовательский центр в сфере искусственного интеллекта НГУ, Россия, 630090, г. Новосибирск, ул. Пирогова, д. 1.*

[2] *Новосибирский государственный технический университет, Россия, 630073, г. Новосибирск, проспект Карла Маркса, 20.*

**Аннотация.** В статье обсуждается подход к построению систем искусственного интеллекта для задач промышленной автоматизации и создания автоматизированных систем управления технологическими процессами в промышленности, городском хозяйстве и на инфраструктурных объектах, прежде всего — на объектах критической инфраструктуры. Предлагается интерпретируемый подход, основанный на иерархии конечных автоматов, оперирующих в пространстве последовательностей состояний объекта управления, где могут быть выявлены причинно-следственные (каузальные) связи на основе причинности по Грейнджеру, между переменными состояния и управляющими воздействиями на систему, представляемые в виде графов переходов между состояниями и их последовательностями. Демонстрируется принципиальная возможность решения данной проблемы на примере симуляции на макете «Умного Дома» на платформе HomeAssistant, а также на реальных полевых данных объекта тепло- и водоснабжения. Данный подход предлагается как основа для построения систем доверенного искусственного интеллекта.

**Ключевые слова:** промышленная автоматизация, система автоматизированного управления технологическими процессами, интерпретируемый искусственный интеллект, доверенный искусственный интеллект, конечный автомат, пространство состояний, граф переходов между состояниями, причинность по Грейнджеру.

**Благодарности:** Результаты получены с использованием ресурсов проектов «Aigents», «Юнисофт Плюс», и «ТехноАналитика».

## *1. Введение*

Развитие систем промышленной автоматизации характеризуется последовательным переходом от жёстко заданных алгоритмов регулирования к адаптивным моделям управления, способным функционировать в условиях неопределённости, неполноты наблюдаемых данных и изменения внешних возмущений. Применение методов машинного обучения в задачах прогнозного обслуживания оборудования, диагностики технологических отклонений, оптимизации энергопотребления и диспетчеризации коммунальной инфраструктуры позволяет повысить качество управления объектами, для которых построение точных аналитических моделей затруднено или экономически неоправданно. Особую значимость данные задачи приобретают применительно к объектам критической инфраструктуры — системам тепло- и водоснабжения, энергоснабжения, транспорта и городского хозяйства, где ошибка управляющего воздействия непосредственно влияет на безопасность и непрерывность предоставления услуг.

Вместе с тем, широкое внедрение моделей, обладающих высокой выразительной способностью (глубоких нейронных сетей, ансамблей деревьев решений, трансформерных архитектур), сопровождается ростом сложности их анализа и верификации. Возникающее противоречие между точностью прогноза и прозрачностью механизма принятия решений описывается в литературе как проблема интерпретируемости искусственного интеллекта (ИИ) [1]. Методы объяснительного интеллекта, в том числе основанные на оценке вклада признаков [2], позволяют охарактеризовать ассоциативные зависимости, зафиксированные моделью, однако не раскрывают структурной организации процесса управления и не гарантируют ни корректности объяснения по отношению к внутренней реализации модели, ни его соответствия физике объекта. Для областей, где требуется обоснование каждого управляющего воздействия, данного уровня обоснованности недостаточно [3].

Дополнительным ограничением выступает нормативный контекст. Стандарты функциональной безопасности (IEC 61508), требования к управлению рисками при применении систем искусственного интеллекта (ISO/IEC 23894), а также положения европейского регулирования в области ИИ предполагают прослеживаемость решений, возможность их независимого аудита и сохранение управляющей роли человека. Отсюда следует, что интерпретируемость в промышленном применении не может рассматриваться как дополнительная характеристика интерфейса пользователя: она должна быть свойством самой модели представления знаний и способа формирования управляющих воздействий. Другими словами, «доверенные» решения на основе ИИ должны обеспечивать не только «локальную интерпретируемость», позволяющую объяснять конкретные решения постфактум, но и «глобальную интерпретируемость», обеспечивающую возможность аудита и экспертной верификации самой модели до начала ее промышленного применения [4].

Перспективный подход к обеспечению доверия реализуется в классах моделей, обладающих встроенной (intrinsically interpretable) семантикой. К их числу относятся конечные автоматы (КА), графы переходов, модели на основе конечных последовательностей состояний и формальные спецификации поведения. Такие модели допускают исчерпывающую трассировку вычислений, проверку инвариантов, воспроизведение сценариев и анализ достижимости нежелательных состояний. Однако их традиционное применение в автоматизированных системах управления технологическими процессами (АСУ ТП) связано с ручной формализацией экспертных знаний, что делает процесс построения модели трудоёмким и плохо масштабируемым на объекты с большим числом переменных состояния.

Перспективным направлением разрешения указанного противоречия является автоматическое извлечение структур представлений из наблюдаемых данных с сохранением формальной семантики. В этом отношении особый интерес представляет анализ временных рядов переменных состояния и управляющих воздействий средствами проверки причинности по Грейнджеру [5]. Необходимо подчеркнуть, что причинность по Грейнджеру выражает не философскую причинную связь, а предсказательный вклад одного сигнала в прогнозирование другого при условии учёта остальной информации. Поэтому обнаружение таких связей в наблюдаемых данных должно сопровождаться проверкой на ложные ассоциации, вызванные скрытыми переменными, общими внешними возмущениями и запаздываниями, а также — по возможности — интервенционным подтверждением в среде симуляции.

В настоящей работе предлагается подход к построению интерпретируемой системы искусственного интеллекта для промышленной автоматизации, в которой объектом моделирования выступает не отдельное наблюдение, а последовательность состояний управляемой системы. Представление может быть реализовано в виде иерархии конечных автоматов, оперирующих в пространстве последовательностей состояний; связи между переменными состояния и управляющими воздействиями выявляются на основании критерия причинности по Грейнджеру и фиксируются в форме графов переходов. Иерархическая организация позволяет разграничить уровни локального поведения, координированного поведения подсистем и целевого управления объектом в целом, обеспечивая согласование формальной проверяемости нижних уровней с адаптивностью уровней верхних.

Принципиальная реализуемость предложенного подхода проверяется на макете «Умного дома» на программной платформе Home Assistant, воспроизводящем распределённое взаимодействие датчиков и исполнительных устройств; а также на реальных полевых данных объекта тепло-водоснабжения, содержащих пропуски, шум измерений и нерегулярность выборки. Совместное рассмотрение этих сред может позволить отделить свойства метода, обусловленные качеством данных, от его структурных ограничений.

Основной вклад работы заключается в: (1) определении процедуры построения иерархии конечных автоматов на основе выявленных каузальных связей; (2) демонстрации применимости подхода к симуляционным и реальным промышленным данным. Совокупность перечисленных свойств — прослеживаемость решения до конкретной последовательности переходов, воспроизводимость, возможность верификации инвариантов и сохранение контура человеческого контроля — рассматривается как основа для построения систем доверенного искусственного интеллекта.

Структура статьи следующая. Раздел 2 содержит обоснование подхода к интерпретируемому моделированию динамических объектов и формулировку требований. Раздел 3 описывает подход к построению иерархии конечных автоматов и процедуру выявления причинных связей. Раздел 4 посвящён экспериментальной апробации предлагаемого подхода для двух сред - «Умного дома» и объекта тепло-водоснабжения. Раздел 5 содержит обсуждение области применимости и возможных ограничений, а раздел 6 — заключение и направления дальнейших исследований.

## 2. Обоснование подхода и формулировка требований

Перспективная, с нашей точки зрения, постановка задачи приводится в работе, где поведение системы психики живого организма либо искусственной системы управления некоторым технологическим объектом может быть формализовано в пространстве состояний, включающих ощущения (сигналы с поля), действия (управляющие воздействия) и ощущения (целевые параметры) [6]. В этом случае, модель поведения организма или управления технологическим процессом может быть рассмотрена как граф переходов между предыдущими последовательностями состояний новых состояний, где каждый переход может характеризоваться его вероятностью, его полезностью с точки зрения потребностей или достижения целевых параметров системы и набором действий, необходимых для реализации данного перехода. По существу, такая модель может быть описана в терминах теории конечных автоматов [7,8].

Имеющийся опыт показывает, что указанный подход может быть успешно применен, в том числе, к задачам обучения класса обучения с подкреплением, в том числе — с отложенным подкреплением [9]. Согласно вычислительной модели, используемой в предшествующих работах [6,9], как показано на рис. 1, агент фиксирует состояния в виде векторов переменных состояния $s_t$ на основе взаимодействия с внешний средой и своих внутренних процессов в каждый момент времени $t$, причем $s_t$ состоит из измерений датчиков $f_t$, степени удовлетворения внутренних потребностей или достижения целевых параметров $y_t$, и фиксации совершенных действий или осуществленных управляющих воздействий $a_t$. Агент в процессе взаимодействия с операционной средой формирует эпизодическую память в виде последовательности векторов состояний как последовательности $T$ состояний, на основе которой могут формироваться модели взаимодействия со средой в виде графов переходов от последовательностей состояний к новым сосояниям $\{s_\tau\}_{\tau \in \{-T,-T+1,\ldots,-1\}} \Rightarrow s_{\tau=0}$, где $\tau$ это относительное время для каждого такого перехода от конечного состояния до одного из предшествующих. Каждый переход из последовательности состояний в новое состояние может в формируемой в ходе обучения модели может характеризоваться значением полезности $U$, характеризующей степень удовлетворения потребностей агента либо степень достижения его текущих целевых параметровв, а также вероятностью перехода в новое состояние $P$, либо счетчик числа данных переходов $C$, позволяющей эту вероятность рассчитать.

Функция обучения $L$ позволяет рассчитать инкрементальное изменение полезности $U$ при переходах между последовательностями состояний в момент времени $t-1$ к новым в момент времени $t$ согласно следующей обобщенной формуле.

$$dU(\{s_\tau\}_{\tau \in \{-T,-T+1,\ldots,-1\}}, s_{\tau=0}) = L(x \cdot (y_t - y_{t-1}), E(a_{t-1}))$$

В приведенной выше формуле $x$ это вектор важности или приоритетности текущих потребностей или целевых параметров, $y_t - y_{t-1}$ это вектор удовлетворенности потребностей системы либо близость ее к заданным целевым параметрам, которая может быть интерпретирована как «награда» с позиций обучения с подкреплением, либо положительной или отрицательной обратной связи в контексте [6,9]. $E(a_{t-1})$ здесь определяет меру затрат энергии или других ресурсов на совершение данного перехода, если предполагается максимизация энергетической или вообще ресурсной (например — экономической) эффективности поведения системы.

Функция принятия решения для такого агента, может быть основана на максимизации комбинации полезности U перехода от последовательности состояний $\{s_\tau\}_{\tau \in \{-T,-T+1,\ldots,-1\}}$ во время $t\text{-}1$ в новое состояние $s_t$ в момент времени $t$ и вероятности P данного перехода, рассчитанной исходя из распределения известных значений счетчиков переходов C для всех известных переходов из исходной последовательности состояннний.

$$s'_t = argmax_s (U(\{s_\tau\}_{\tau \in \{-T,-T+1,\ldots,-1\}}, s_{\tau=0}), P(\{s_\tau\}_{\tau \in \{-T,-T+1,\ldots,-1\}}, s_{\tau=0}))$$

В то время как данный подход может быть реализован в парадигме теории конечных автоматов [7,8], известной проблемой его применения является комбинаторный взрыв состояний в плоских автоматах при моделировании сложных систем [10,11], что в нашем случае усугубляется тем, что фактически состояние автомата определяется целой последовательностью состояний $s$ за некоторый интервал времени $T$. Для эффективного решения данной проблемы требуется как повышение дискретности значений каждой переменной состояния, так и выделение в однородном пространстве состояний s сегментов или иерархий, соответствующих различным подсистемам самого агента и управляемой им среды, с пониженной размерностью для каждого такого сегмента или раздела иерархии. Это должно позволить как понизить число возможных состояний каждой отдельно взятой переменной, так и минимизировать число переменных в каждом отдельно взятом потоке состояний, соответствующих отдельному конечному автомату в иерархии таких автоматов.

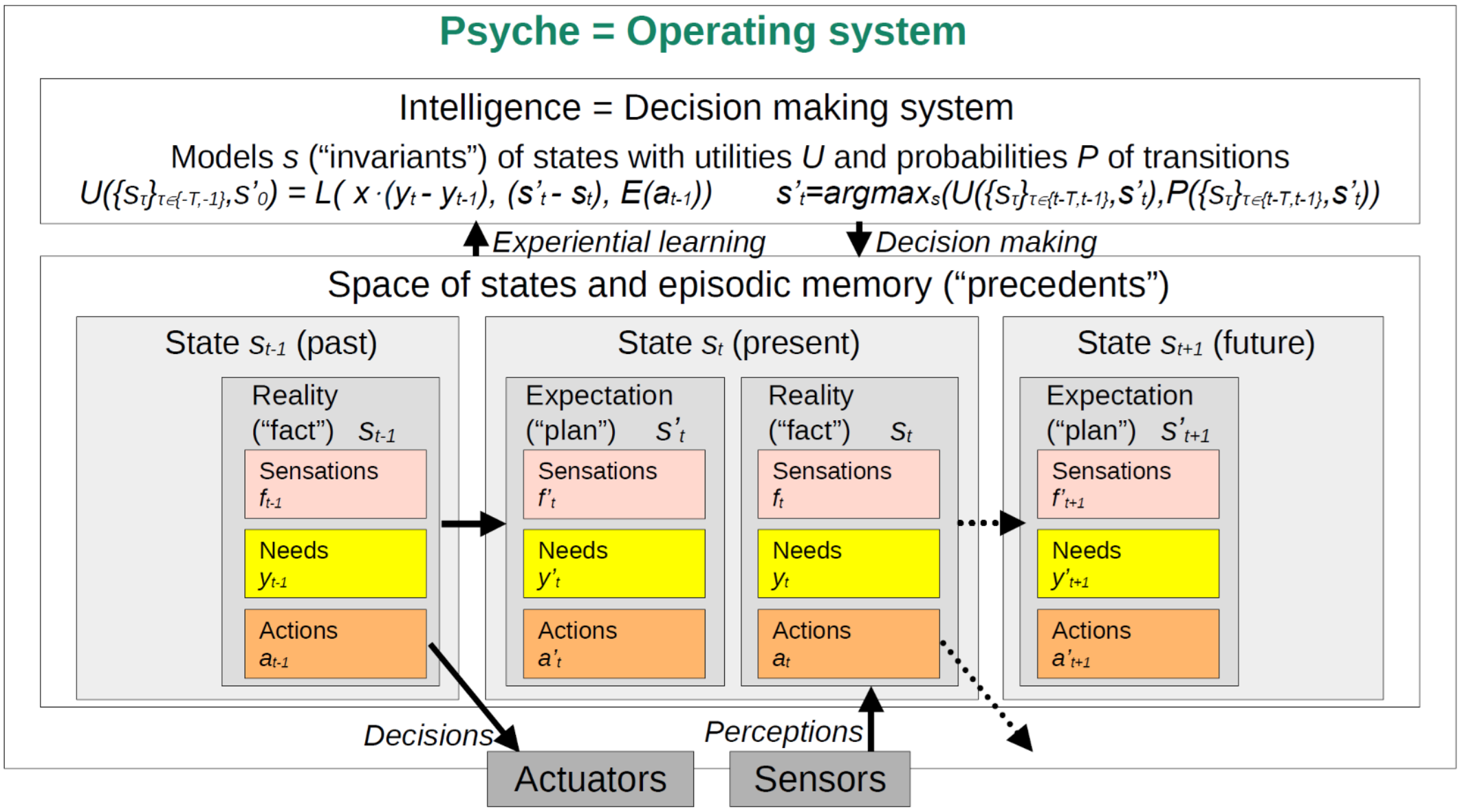

*Рис. 1. Иллюстрация процесса принятия решений в пространстве состояний, основанного на графах переходов между состояниями, которые хранятся в эпизодической памяти агента. В этой модели наблюдавшиеся ранее состояния s представляют собой эпизоды памяти агента — своего рода прецеденты, связанные между собой переходами. С каждым таким переходом соотнесены мера его полезности и количество зафиксированных случаев (прецедентов), в то время как прогнозируемые состояния s' — это ожидания, возникающие у агента в любой конкретный момент: будь то процесс принятия решения или пассивное наблюдение за окружающей средой.*

## *3. Подходы к построению модели*

Проблема комбинаторного взрыва числа состояний в плоских автоматах [7,8] решается как повышением дискретности каждой переменной состояния, так и понижением числа переменных состояния — разделением общего пространства переменных на изолированные группы (соответствующие отдельным физическим подсистемам объекта управления) или пересекающиеся группы (в случае, если какие-то переменные среды влияют на несколько группы или повержены влиянию нескольких групп). Соответствующие выделенные группы переменных могут быть приведены в соответствие отдельным конечным автоматам, и модель управления объектом (модель поведения агента управления) может быть сведена к суперпозиции или иерархии таких автоматов.

### 3.1 Квантизация сигналов

Квантизация сигналов предполагает оценку характера распределения возможных значений переменной, соответствующей сигналу, с целью оценки характера распределения значений, выявление наличия нескольких модальностей, потенциально соответствующих различным состояниям или режимам системы, параметр которой определятся сигналом, значения которого представлены соответствующей переменной.

В нашей предшествующей работе [12] решение этой задачи предполагается комбинацией метрики «dip test» (статистический тест Хартигана) [13], определяющего степень одно-модальности распределения и коэффициента «силуэта» («Silhouette Coefficient») [14], позволяющего оценить оптимальность кластеризации при заданном числе кластеров во многомерном пространстве (в нашем случае, при кластеризации методом K-means [12] в одно-мерном пространстве).

В данной работе мы использовали тот-же самый подход — оценивали возможность выявления более одной модальности по метрике «dip test» и определяли оптимально число модальностей как число кластеров максимизирующих значение коэффициента «силуэта» при одно-мерной кластеризации K-means. Соотвественно, квантизация осуществлялась заменой исходных значений сигнала, его значениями, соответствующими «центроиду» ближайшего «кластера» K-means при оптимальном числе кластеров K, соответствующем числу мод или режимов сигнала.

## 3.2 Выявление групповых состояний

Задача выявления групп сигналов, представляющих параметры среды или системы, взаимосвязанные в силу их внутренней природы, рассматривалась нами в предшествующих работах по каузальной и предиктивной аналитике применительно к анализу финансовых временных рядов [15,16]. Для этого использовался анализ значений коэффициента корреляции по Пирсону между значениями временных рядов, с получением матриц взаимной корреляции, со значениями, указывающими степень взаимозависимости сигналов прямой (положительные значения) или обратной (отрицательные значения). Также — для выявления потенциальных причинно-следственных связей на различных временных горизонтах на основе принципа каузальности по Грейнджеру [5], анализировались значения коэффициентов корреляции по Пирсону между различными временными рядами с различными временными сдвигами, с определением таких сдвигов, которые максимизируют абсолютное значение коэффициента корреляции (отрицательной либо положительной связи). Выявленные таким образом потенциальные причинно-следственные связи использовались для построения предиктивных моделей, основанных, для каждого целевого сигнала, на сигналах, обладающих наибольшей выявленной каузальной связью по отношению к целевому.

В ходе представляемого нами исследования целью было выявление отдельных групп сигналов, связанных между собой на различных временных горизонтах, в связи с чем мы исследовали возможность кластеризации сигналов всего объекта управления в группы, соответствующие отдельным подсистемам объекта. Для этих групп уже можно было строить предиктивные модели в пространстве состояний этих подсистем, с целью управления системой в целом на основе иерархии конечных автоматов, отражающей иерархию подсистем объекта управления.

В исследовании мы обнаружили, что многие сигналы характеризуются значимыми изменениям уровня сигнала на протяжении периода наблюдений, то есть — низкочастотный изменения (в масштабе месяцев и дней) являются существенно более значимыми по амплитуде сигнала, чем высокочастотные изменения в масштабе часов и минут. В связи с этим, анализ взаимной корреляции между сигналами как «в моменте» (без временного сдвига), так и с временными сдвигами (с целью каузальной аналитики по Грейнджеру) осуществлялся как на исходных сигналах (с выявлением низкочастотных связей на масштабах месяцев и дней), так и на приведенных к общему уровню высокочастотных данных (часы и минуты), полученных нормализаций сигналов $S_t = S_t / max(abs(S_t))$ и вычитанием низкочастотного «тренда» $S^*_t = S_t - rollingmean_{window=7}(S_t)$, полученного сглаживанием в скользящем окне (например, для второй опытной среды ширина окна для часовых измерений опытным путем была определена как 6 часов, или семь точек измерения). Анализ распределения значений матриц взаимной корреляции позволил выделить кластеризацией группы сильно взаимосвязанных сигналов с коэффициентами взаимной корреляции (для второй опытной среды выше 0.9), а анализ каузальности по Грейнджеру позволил выделить пары сигналов, потенциально связанных причинно-следственными связями с различными временными горизонтами (для второй опытной среды — 1 час и 6 часов).

Выделенные, как описано выше, парные связи позволили строить группы взаимосвязанных сигналов на различных временных горизонтах. Например, для второй опытной среды можно было выявить группы: а) связанные «в моменте» (на основе коэффициентов корреляции без сдвига), б) связанные причинно-следственными связями на часовом интервале, в) связанные причинно-следственными связями на шести-часовом интервале.

## *4. Экспериментальные результаты*

### 4.1 «Умный дом»

Первой исследованной средой была симуляция проекта автоматизации «Умного дома», выполненная на основе платформы «Home Assistant» (https://www.home-assistant.io/). На основе этой платформы была развернута виртуальная конфигурация двух комнат с датчиками (сенсорами) движения и осветительным оборудованием (лампами), как показано на рис. 2. В исследовании ставилась задача автоматического формирования модели переходов состояний, позволяющей адаптивно подстраивать автоматическое включение и выключение осветительного оборудования в режиме реального времени на основе накопления статистики о поведении виртуального обитателя дома, фиксируемого сенсорами движения и включением либо выключением осветительного оборудования.

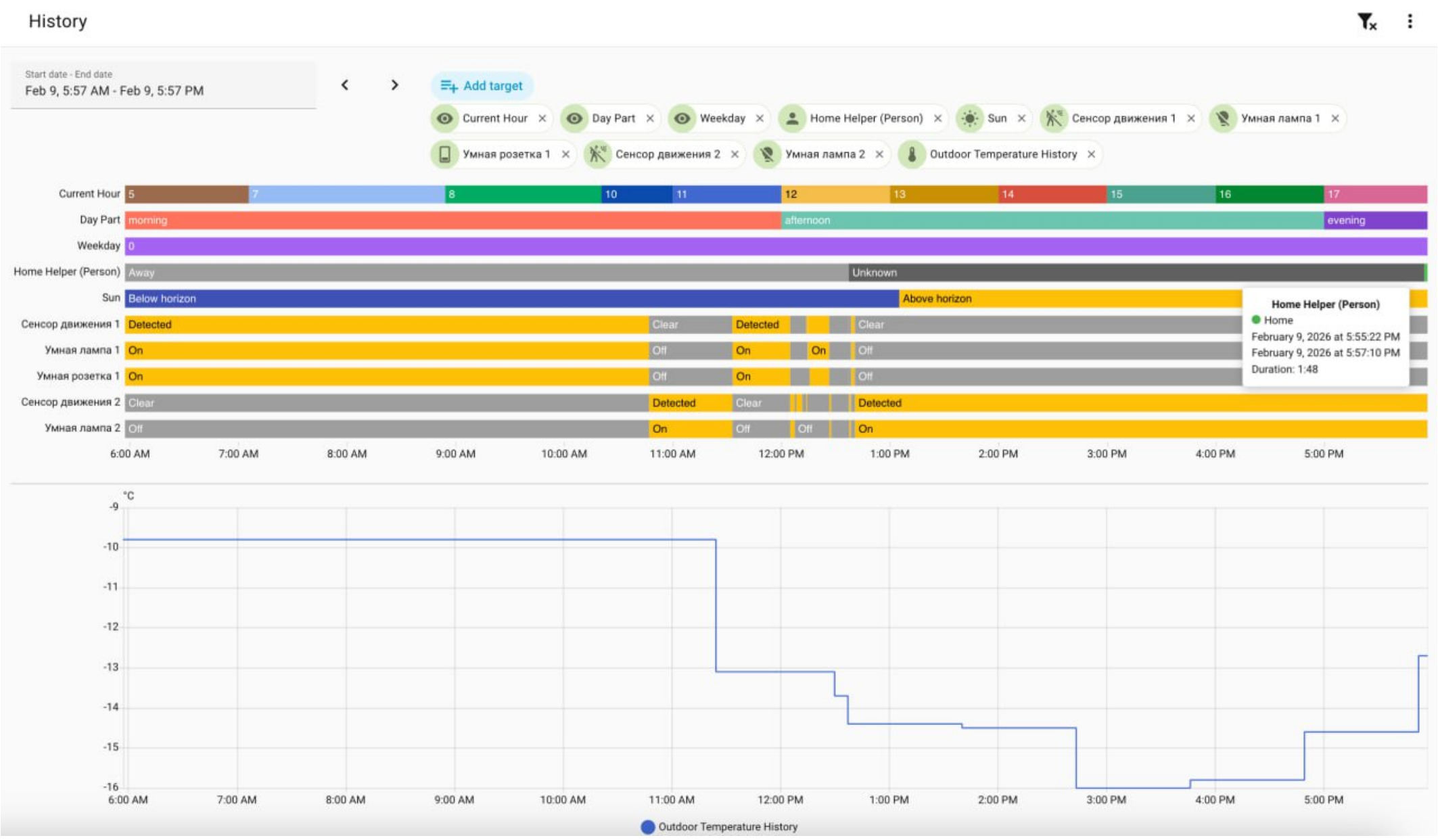

*Рис. 2. Визуализация временной динамики измеряемых и управляющих сигналов виртуальной конфигурация двух комнат с датчиками (сенсорами) движения и осветительным оборудованием (лампами) на платформе Home Assistant.*

По итогам исследования были выделены выраженные паттерны или состояний совместной активизации датчиков движения и определенных комбинаций включенных осветительных приборов. При достаточной статистической выраженности паттернов стало возможным предсказание переходов таким образом, что частичная активация некоего состояния (например, срабатывания датчика движения в одной комнате вместе с включением одной иза ламп в этой же комнате) может автоматически приводить активацию управляющих сигналов, доводящих текущее состояние до полного сочетания со статистически ожидаемым (например, включением недостающих ламп, которые обычно являются включенными в этой ситуации).

## 4.2 Система тепло- и водо- снабжения городского хозяйства

Объектом исследования являлся участок двухконтурной системы теплоснабжения с промежуточными теплообменными пунктами (рис. 3).

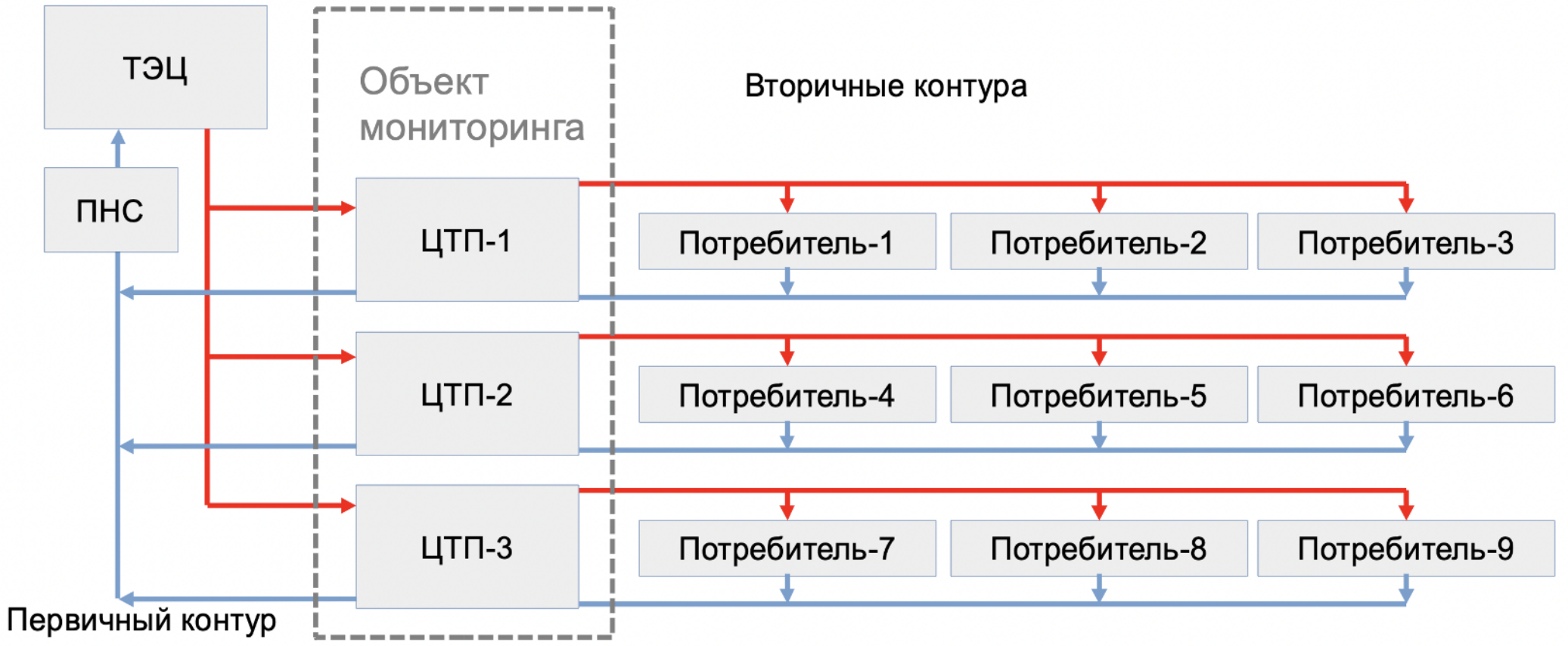

*Рис. 3. Принципиальная схема двухконтурной системы теплоснабжения с тремя ЦТП, ПНС-1 и контуром ГВС (потребители указаны условно).*

Первичный контур образован магистралями от теплоэлектроцентрали (ТЭЦ) через три центральных тепловых пункта (ЦТП-1, ЦТП-2, ЦТП-3) и повысительную насосную станцию обратного трубопровода ПНС с возвратом теплоносителя на ТЭЦ. К каждому ЦТП присоединено по пять потребителей во вторичном контуре, всего 15. Контур горячего водоснабжения (ГВС) включает подогреватели ГВС в ЦТП; холодная вода подаётся на ЦТП от городского водопровода. Подпитка первичного контура осуществляется в ЦТП и измеряется расходомером М94. Существенной особенностью объекта является геодезический перепад: ТЭЦ расположена на возвышенности (плато), абоненты — в низине. Разность отметок обусловливает наличие ПНС на обратном трубопроводе.

Границы измерений (рис. 3) охватывают первичный контур на участке от входов ЦТП до ПНС, контуры ГВС и вводы холодной воды. Отпуск тепловой энергии с ТЭЦ, параметры на напорной стороне ПНС и параметры у абонентов в состав данных не входят. Следствием является незамкнутость теплового баланса магистрали «плато — низина»: количество тепловой энергии, отпущенной источником, не может быть сопоставлено с суммой Q по ЦТП. Баланс проверяется только по отдельным узлам — по невязке расходов первичного контура и по соотношению расходов контура ГВС. Данные представляют собой архив телеметрии трёх ЦТП за период с 10.01.2021 по 09.12.2024. Исходная выборка нерегулярна; для анализа она приведена к равномерной часовой сетке усреднением, при этом интервалы без данных сохраняют признак пропуска. Состав и характеристики каналов приведены в табл. 1. Набор каналов неодинаков для разных ЦТП: часть каналов в архиве отдельных узлов отсутствует. При построении графа связей такой канал исключался из соответствующей компоненты, а не восстанавливается по соседним узлам: восстановление внесло бы в анализ связь, которой в данных нет. Также, для получения приведенных ниже результатов были исключены записи за рамками периода с 13.12.2022 по 16.04.2023 ввиду наличия за его рамками значительных участков с отсутствующими либо константными значениями по тем или иным сигналам. В итоге представляемые результаты были получены по 2106 измерениям для каждого из 55 сигналов.

*Табл. 1. Каналы измерений объекта.*

| Обозн. | Физический смысл | Контур | Ед. изм. |
|---|---|---|---|
| T1 | Температура теплоносителя в подающем трубопроводе | первичный | °C |
| T2 | Температура теплоносителя в обратном трубопроводе | первичный | °C |
| P1 | Давление в подающем трубопроводе | первичный | кгс/см$^2$ |
| P2 | Давление в обратном трубопроводе | первичный | кгс/см$^2$ |
| M1 | Массовый расход на входе ЦТП | первичный | т/ч |
| M2 | Массовый расход на выходе ЦТП | первичный | т/ч |
| M94 | Расход подпитки | первичный | т/ч |
| T3 | Температура горячей воды в подающем трубопроводе | ГВС | °C |
| T4 | Температура горячей воды в обратном трубопроводе | ГВС | °C |
| P3 | Давление в подающем трубопроводе ГВС | ГВС | кгс/см$^2$ |
| P4 | Давление в обратном трубопроводе ГВС | ГВС | кгс/см$^2$ |
| M3 | Массовый расход ГВС на входе | ГВС | т/ч |
| M4 | Массовый расход ГВС на выходе | ГВС | т/ч |
| $T_{хвс}$ | Температура холодной воды на вводе ЦТП | ХВС | °C |
| $P_{хвс}$ | Давление холодной воды на вводе ЦТП | ХВС | кгс/см$^2$ |
| $V_{хвс}$ | Объёмный расход холодной воды | ХВС | м$^3$/ч |
| $V_{хвсб}$ | Объёмный расход по байпасу | ХВС | м$^3$/ч |
| Q | Тепловая энергия отопления и вентиляции | первичный | Гкал/ч |
| $Q_{гвс}$ | Тепловая энергия ГВС | ГВС | Гкал/ч |
| $T_{нв}$ | Температура наружного воздуха | внешняя | °C |

В ходе исследований по квантизации было выявлено два вида сигналов — характеризующихся около-нормальным распределением значений («dip test» > 0.65 и «Silhouette Coefficient» < 0.8) и характеризующихся около-мультимодальным распределением («dip test» <= 0.65 и «Silhouette Coefficient» >= 0.8). Для последних сигналов типичное число мод, выявленных анализом составляло 2-3. Для решения последующей задачи выявления конечных состояний для построения предиктивных моделей на основе конечных автоматов ко всем сигналам применялась квантизация соотвественно оптимальному числу мод от 2 до 3, в зависимости от максимального значения «Silhouette Coefficient» для данного числа мод, с заменой каждого занчения сигнала на максимально близкое к нему значение соответствующей моды, вычисленное как среднее для моды (соответствующее центроиду кластера одномерной кластеризации). Пример квантованных сигналов приведен на рис. 4.

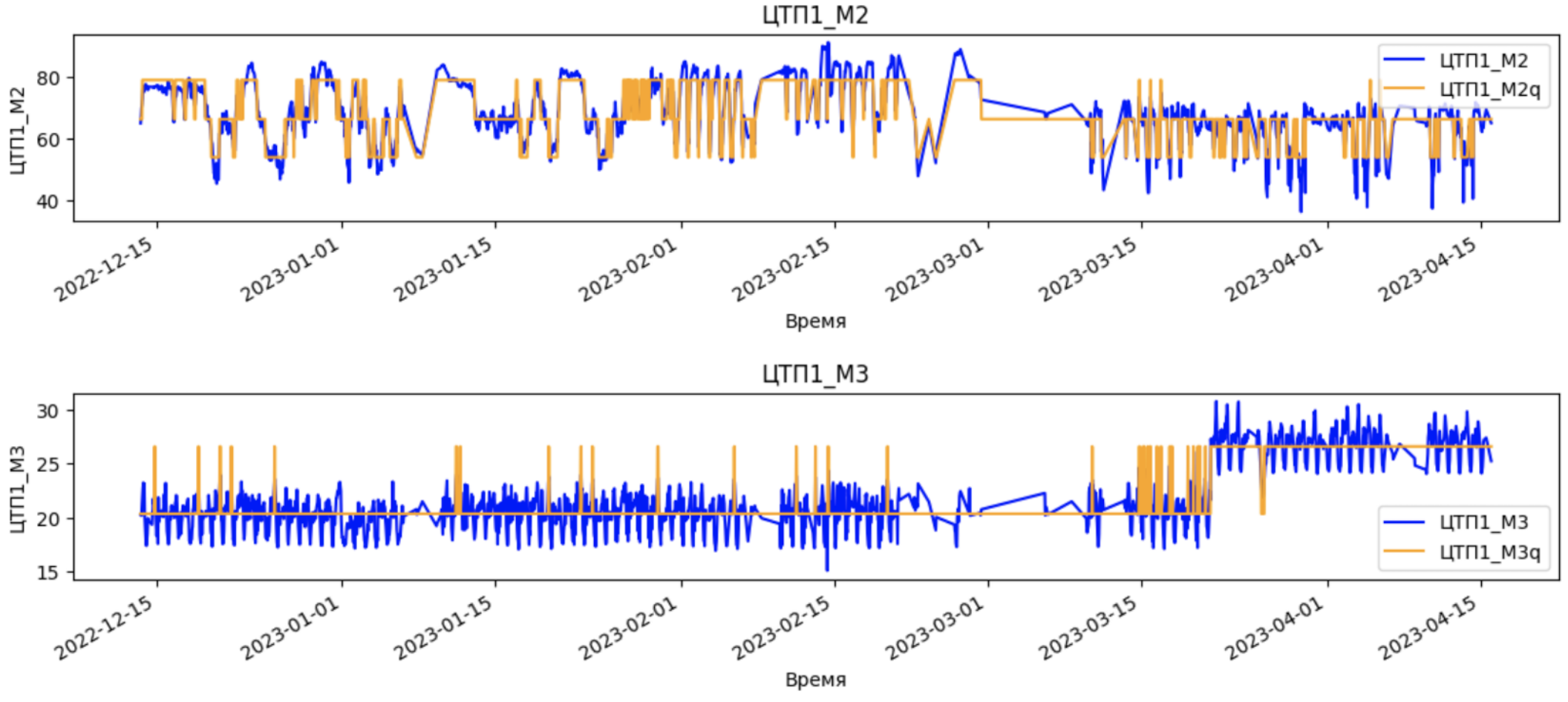

*Рис. 4. Результаты квантизации для двух сигналов (синий — исходный сигнал, оранжевый — квантизованный).*

Корреляционный анализ «в моменте» применительно к исходным сигналам и сигналам с удаленным низкочастотным «трендом» позволил сформировать два вида групп сигналов (рис. 5).

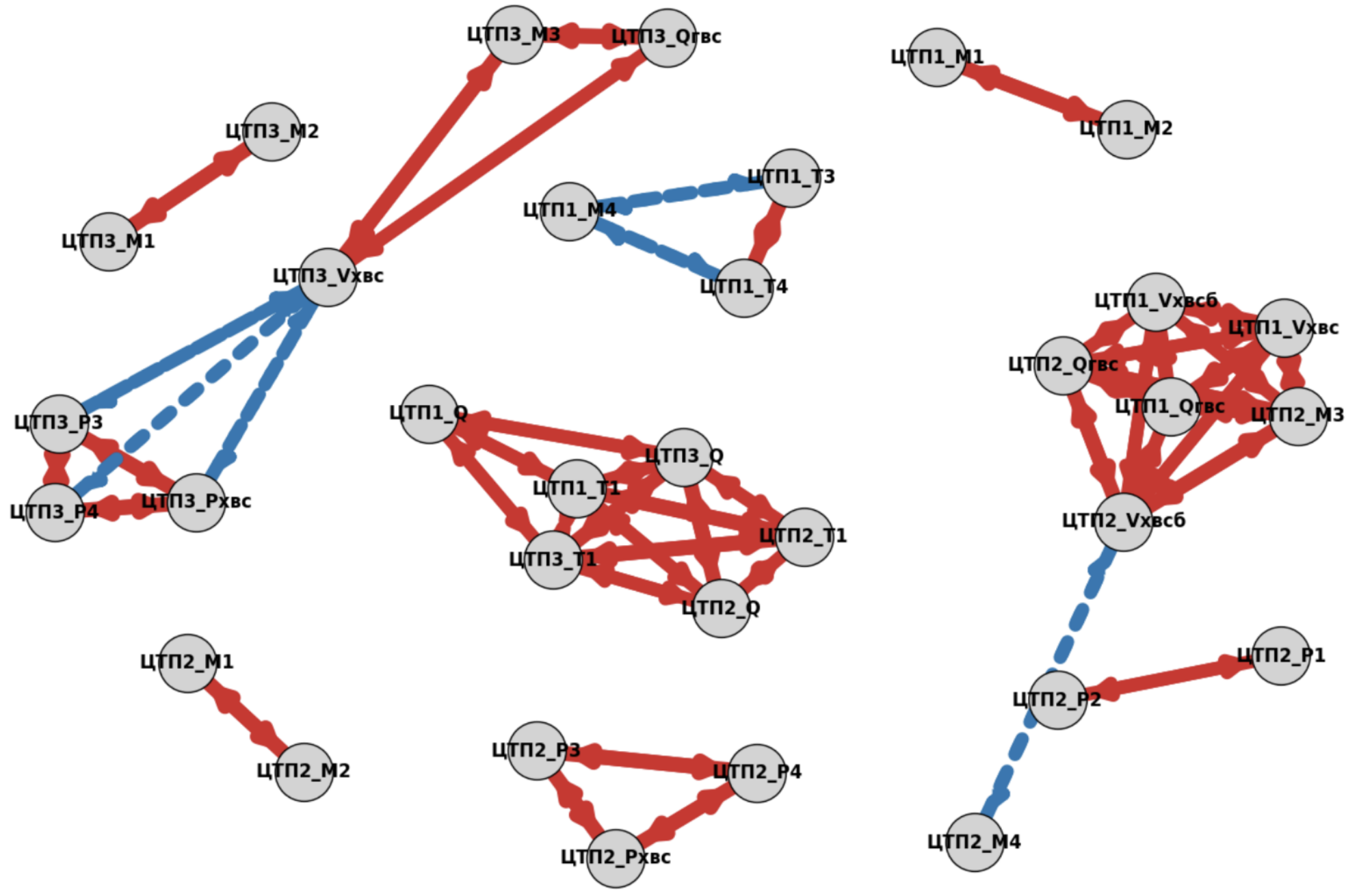

*Рис. 5. Граф связности сигналов на основе положительной (красные сплошные стрелки) и отрицательной (синие пунктирные стрелки) взаимной корреляции по Пирсону с абсолютными значениями выше 0.9; толщина стрелок пропорциональна значению коэффициента корреляции.*

В первом случае это оказались группы сигналов, связанные низкочастотной корреляцией (на уровне дней и месяцев), имеющих общие сезонные составляющие либо относящихся к подсистемам объекта, имеющим общие условия в том или ином режиме функционирования (например, общая аварийная ситуация либо общий штатный режим). Во втором случае выявлялись группы относительно высокочастотных сигналов с удаленным «трендом», рассчитанным как скользящее среднее по 7 часам. На рис. 5 показаны группы исходных сигналов с высокой степенью статистической связанности - прямой либо обратной корреляции с абсолютным значением коэффициента Пирсона более 0.9.

Полученный граф распадается на компоненты, соответствующие физически различимым подсистемам: первичный контур отдельного ЦТП — соотношение расходов, перепада давлений и температур; общий температурный режим — согласованное изменение T1 на разных ЦТП с транспортным запаздыванием; контур ГВС — баланс расходов M3, $V_{хвс}$ и тепловой энергии $Q_{гвс}$; гидравлика ГВС — давления P3, P4 и $P_{хвс}$. Связи между одноимёнными каналами разных ЦТП сохраняются в графе только при наличии ненулевого лага, согласующегося с транспортным запаздыванием в магистрали. При нулевом лаге они интерпретируются как проявление общего внешнего фактора и из графа удаляются.

Выявление причинности по Грейнджеру выполнялось после вычитания из сигнала «тренда», включающего сезонную составляющую в пределах окон однородного режима, что обеспечивало приблизительную стационарность рядов (рис. 6).

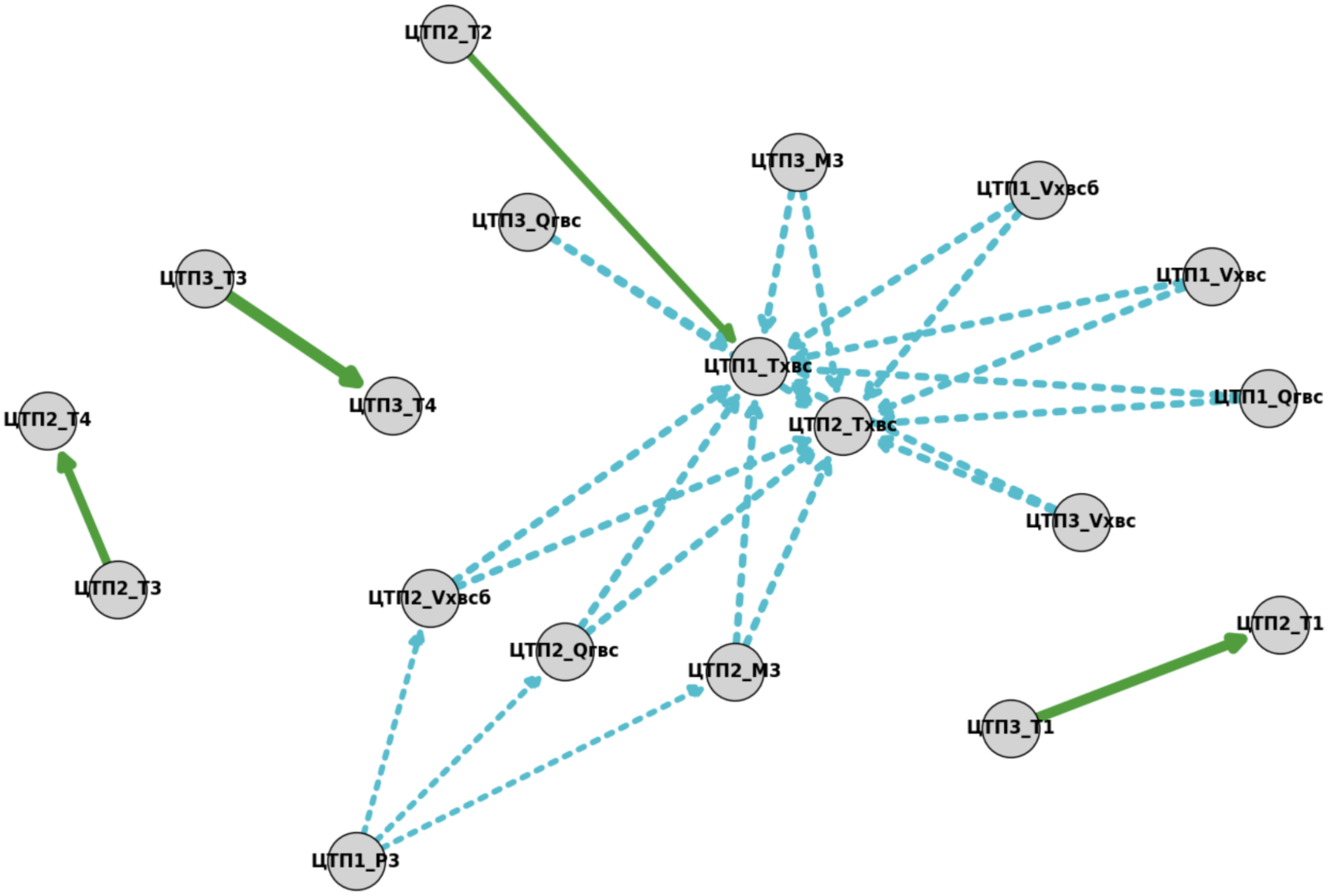

*Рис. 6. Граф потенциальной причинной связности сигналов (по Грейнджеру) на основе положительной либо отрицательной и отрицательной взаимной корреляции с различными временными сдвигами: зеленые сплошные стрелки — 1 час, голубые пунктирные стрелки — 6 часов; толщина стрелок пропорциональна значению коэффициента корреляции.*

Для удаления «тренда», экспериментально была определена ширина скользящего окна 6 часов (7 точек измерения) для расчета вычитаемого тренда. Поиск временных сдвигов, дающих максимальное значение коэффициента корреляции осуществлялся для каждой пары сигналов со сдвигами от 1 до 24 часов. Результаты кластеризации сигналов с выявленными потенциально причинно-следственными связями показаны на рис. 6. При этом были выявлены два класса связей — относительно краткосрочные связи со сдвигом 1 час и относительно долгосрочные — со сдвигом 6 часов. Это позволяет говорить о возможности построения интерпретируемых предиктивных моделей для ряда параметров на основе обнаруженных гипотетических причинно-следственных связей между фиксируемыми сигналами для 6-часовых и 1-часовых горизонтах прогноза.

Существенное ограничение метода для данного объекта состоит в том, что контуры теплоснабжения замкнуты регуляторами: температура ГВС и перепад давлений поддерживаются автоматическими регуляторами, вследствие чего измеряемые ряды формируются в контуре обратной связи «уставка — регулирующий клапан — расход — температура». В таких условиях критерий выявляет направленный вклад в предсказуемость, а не физическую причинность, и может указывать направление, противоположное фактическому потоку воздействия. Результаты интерпретируются как «направленное влияние в смысле предсказуемости» и формулируются в контексте работы именно так.

Отдельно проверяется устойчивость результата к способу удаления сезонности. Связи, сохраняющиеся при переходе от уровней к остаткам, рассматриваются как физически обоснованные; связи, исчезающие при удалении сезонной составляющей, к рассмотрению не принимаются. Так, в предварительном варианте анализа связь между температурой обратной сетевой воды ЦТП-2 и температурой холодной воды ЦТП-1 имела высокий уровень значимости на уровнях рядов, но исчезала после удаления сезонной составляющей; она классифицирована как артефакт общего внешнего фактора. Единственная связь между объектами, устойчивая ко всем вариантам обработки, — направленное влияние температуры сетевой воды ЦТП-3 на температуру ЦТП-2.

Эксперимент по построению модели конечных автоматов на основе полученных квантованных сигналов первоначально был проведен без выделения групп сигналов, на всех 55 сигналах для 2106 измерений по каждому из каналов. В этом случае было обнаружено всего 1968 уникальных состояний, то есть только менее 10% всех состояний оказались воспроизводимыми всего по несколько раз за исследуемый период, что указало на практическую невозможность построить модель всей системы в целом для объекта управления.

Далее, этот эксперимент был повторен уже индивидуально для каждой из выделенных выше на основе взаимной корреляции (рис. 5) групп сигналов и в этом случае число уникальных состояний оказалось на три порядка меньше, как показано в табл. 2. То есть, по каждой из выявленных групп, подавляющее число зафиксированных дискретных состояний по многоканальным группам квантованных сигналов оказалось представленным множеством эпизодов, что может позволить строить предиктивные модели на основе теории конечных автоматов и графов переходов между последовательностями состояний и состояниями.

*Табл. 2. Количества выявленных дискретных состояний по группам квантованных каналов.*

| Переменные состояний в группе сигналов | Число состояний |
|---|---|
| ЦТП1_М1, ЦТП1_М2, ЦТП1_П2 | 15 |
| ЦТП1_М3, ЦТП1_М4, ЦТП1_Т3, ЦТП1_Т4 | 7 |
| ЦТП1_П3, ЦТП1_Qгвс, ЦТП1_Вхвс, ЦТП1_Вхвсб, ЦТП2_М3, ЦТП2_М4, ЦТП2_Qгвс, ЦТП2_Вхвсб, ЦТП3_М3, ЦТП3_П3, ЦТП3_П4, ЦТП3_Пхвс, ЦТП3_Qгвс, ЦТП3_Вхвс | 135 |
| ЦТП1_Q, ЦТП1_Т1, ЦТП1_Т2, ЦТП2_Q, ЦТП2_Т1, ЦТП3_Q, ЦТП3_Т1, ЦТП3_Т2 | 62 |
| ЦТП2_М1, ЦТП2_М2, ЦТП3_П1 | 13 |
| ЦТП2_П1, ЦТП2_П2 | 4 |
| ЦТП2_П3, ЦТП2_П4, ЦТП2_Пхвс | 3 |
| ЦТП3_М1, ЦТП3_М2 | 4 |

## *5. Области применимости и ограничения подхода*

### 5.1 Области применимости

Предлагаемый подход к построению интерпретируемого искусственного интеллекта, основанный на иерархии конечных автоматов и анализе причинно-следственных связей по Грейнджеру в пространстве последовательностей состояний, открывает новые возможности для управления сложными динамическими системами. В отличие от статистических (нейросетевых) моделей «черного ящика», данный метод обеспечивает формальную трассируемость решений, что критически важно для объектов критической инфраструктуры, например, для систем жилищно-коммунального хозяйства (ЖКХ).

Например, применение описанного подхода в ЖКХ трансформирует систему управления из реактивной (реагирующей на уже случившиеся сбои) в проактивную и предиктивную. При этом строгая математическая формализация в виде конечных автоматов гарантирует, что сложность алгоритмов не жертвует понятностью и контролируемостью процессов для человека-оператора.

Одним из основных преимуществ иерархии КА и анализе причинно-следственных связей является снижение размерности. На примере нашего исследования, декомпозиция 55 каналов на восемь групп по 2–14 сигналов (табл.2) сокращает объём ручного анализа: число различимых состояний в группе не превышает 135 против 1968 в исходном варианте. Построенная модель позволяет выявлять и фиксировать в графе переходов наиболее энергоэффективные режимы работы, отсекая не оптимальные. Еще одной важной областью применения является выявление каузальных связей, что может позволить системе рекомендовать не резкие, дискретные изменения параметров, а плавные, своевременные и прогнозируемые смены режимов, что, на нашем примере,  должно снижать гидравлические и тепловые удары в сети.

## 5.2 Ограничения подхода

Основные ограничения подхода в данный момент видятся в недостаточной практической апробации. В частности, в случае тестирования подхода в среде «Умного Дома», среда была чрезмерно упрощенной, фактического выделения групп сигналов не производилось и общая модель строилась для всего объекта в целом на ограниченном временном диапазоне, что нельзя считать убедительным, нужна апробация на реальных физических конфигурациях с построением реальных иерархий подсистем в рамках заявленного подхода.

Применительно к эксперименту описанному для объекта тело- водо- снабжения, остается спорным вопрос, насколько квантизация сигналов с около-нормальным распределением является с выделением всего 2 или 3 дискретных состояний является правомочной. В целом, полученные в результате исследования подсистемы конечных состояний по группам сигналов требуют подтверждения практической значимости — требуется построение реальных предиктивных моделей с оценкой их фактической точности. Также требуют отдельной верификации предиктивная способность выявленных причинно-следственных связей по Грейнджеру, с учетом экспертной валидации всех выявляемых связей.

## *6. Заключение*

В представленной работе мы показали потенциал практической применимости интерпретируемого ИИ на основе теории конечных автоматов в пространстве состояний переменных среды и графов переходов между последовательностями дискретных состояний и новыми состояниями, на основе квантизации исходных сигналов, с построением иерархии потоков состояний, соответствующих иерархий соответствующих конечных автоматов — с помощью корреляционного анализа сигналов и анализа потенциальных причинных связей между сигналами по Грейнджеру.

При дальнейшем развитии работ мы планируем валидацию предложенного подхода на более реалистичных симуляциях и реальных инсталляциях «Умного Дома», а также на более полных, объемных и качественных данных объектов ЖКХ и промышленных объектов, включая полноценную реализацию и апробацию предиктивных моделей, основанных на предложенном подходе.

## Список литературы

## *Информация об авторах*

Антон Германович КОЛОНИН — кандидат технических наук, ведущий научный сотрудник Исследовательского центра в сфере искусственного интеллекта и старший преподаватель ММФ НГУ. Основатель программной платформы с открытым кодом Aigents, ведущий исследователь ИИ в компании SingularityNET и создатель сообщества AGI-in-Russian. Сфера научных интересов: сильный искусственный интеллект (AGI), обработка естественного языка (NLP), архитектуры блокчейн, анализ социальных сетей, извлечение знаний и, машинное обучение и промышленная автоматизация.

Александр Игоревич ДВОРЦЕВОЙ — кандидат технических наук, доцент кафедры тепловых электрических станций НГТУ НЭТИ. Руководитель проектов компании «Юнисофт Плюс» (разработка IIoT-платформы для 518 технологических линий на 198 предприятиях для компаний их списка Fortune 500) и научный руководитель «ТехноАналитика» (резидент Сколково). Сфера научных интересов: автоматизация и цифровое управление теплоэнергетическими объектами, АСУ ТЭС, цифровые двойники, мониторинг и диагностика энергетического оборудования методами машинного обучения (индивидуальные тепловые пункты, силовые трансформаторы), моделирование и оптимизация режимов энергоблоков, эксергетический анализ и перерасход топлива, применение LLM для интерпретации технологических данных.